\pdfoutput=1

\documentclass[11pt]{article}

\usepackage[]{emnlp2021}

\usepackage{times}
\usepackage{latexsym}
\usepackage{amsmath}
\usepackage{amssymb}
\usepackage{hyperref}
\usepackage{graphicx} 
\usepackage{subcaption}
\usepackage{adjustbox}
\usepackage{booktabs}
\usepackage{multirow}
\usepackage{lscape}
\usepackage{xcolor}
\usepackage{colortbl}
\usepackage{xspace}
\usepackage{tabularx}
\usepackage{scalerel}
\usepackage[most]{tcolorbox}
\usepackage[ruled,vlined]{algorithm2e}
\SetKwProg{Init}{initialization}{}{}

\newcommand\blankfootnote[1]{%
  \let\thefootnote\relax\footnotetext{#1}%
  \let\thefootnote\svthefootnote%
}

\definecolor{myblue}{HTML}{2B79B0}
\definecolor{myorange}{HTML}{FB7F36}
\definecolor{mygreen}{HTML}{389E3B}
\definecolor{myred}{HTML}{D22D35}

\newcommand{\ssc}[1]{{\small \sc #1}\xspace}
\newcommand{\bssc}[1]{{\small \sc \textbf{#1}}\xspace}
\newcommand{\fssc}[1]{{\scriptsize \sc #1}\xspace}
\newcommand{\strata}{{\ssc{STraTA}}\xspace}
\newcommand{\lstrata}{{\sc{STraTA}}\xspace}
\newcommand{\bstrata}{{\bssc{STraTA}}\xspace}

\newcommand\myfontsize{\fontsize{10pt}{11pt}\selectfont}
\newcommand{\bmmc}[1]{\bm{\mathcal{#1}}}
\newcommand{\smallsup}[1]{\scaleto{\text{#1}}{4pt}}
\newcommand{\std}[1]{\mbox{\small{\textsc{Std}}}}
\newcommand{\bertlarge}[1]{\ssc{BERT$_{\smallsup{Large}}$}}
\newcommand{\robertalarge}[1]{\ssc{RoBERTa$_{\smallsup{Large}}$}}
\newcommand{\bertbase}[1]{\ssc{BERT$_{\smallsup{Base}}$}}
\newcommand{\randbase}[1]{\ssc{RAND$_{\smallsup{Base}}$}}
\newcommand{\lbertlarge}[1]{\sc{BERT$_{\textsc{Large}}$}}
\newcommand{\lrobertalarge}[1]{\sc{RoBERTa$_{\text{Large}}$}}
\newcommand{\lbertbase}[1]{\sc{BERT$_{\textsc{Base}}$}}
\newcommand{\lrandbase}[1]{\sc{RAND$_{\textsc{Base}}$}}

\usepackage{mdwlist}
\usepackage{latexsym}
\usepackage{nicefrac}
\usepackage{booktabs}
\usepackage{amsfonts}
\usepackage{bold-extra}
\usepackage{amsmath}
\usepackage{dsfont}
\usepackage{amssymb}
\usepackage{bm}
\usepackage{graphicx}
\usepackage{mathtools}
\usepackage{microtype}
\usepackage{multirow}
\usepackage{multicol}
\usepackage{latexsym,comment}

\newcommand{\gem}[1]{\mbox{\textsc{gem}}}

\newcommand{\hidetext}[1]{}
\newcommand{\ignore}[1]{}

\newcommand{\smallurl}[1]{ \begin{tiny}\url{#1}\end{tiny}}

\definecolor{lightblue}{HTML}{3cc7ea}
\definecolor{grey}{rgb}{0.95,0.95,0.95}
\definecolor{ceil}{rgb}{0.57, 0.63, 0.81}

\usepackage[T1]{fontenc}

\usepackage[utf8]{inputenc}

\usepackage{microtype}

\title{STraTA: Self-Training with Task Augmentation \\ for Better Few-shot Learning}

\author{Tu Vu$^{1,2}$$^\bigstar$, Minh-Thang Luong$^{1}$,  Quoc V. Le$^{1}$, Grady Simon$^{1}$, Mohit Iyyer$^{2}$ \\
  Google Research$^1$ \\
  University of Massachusetts Amherst$^2$ \\
  \texttt{\{ttvu,thangluong,qvl,gradys\}@google.com}\\
  \texttt{\{tuvu,miyyer\}@cs.umass.edu}\\}

\date{}

\begin{document}
\maketitle
\begin{abstract}
\label{section:abstract}
Despite their recent successes in tackling many \ssc{NLP} tasks, large-scale pre-trained language models 
do not perform as well in few-shot settings where only a handful of training examples are available.
To address this shortcoming, we propose \bstrata, which stands for \textbf{S}elf-\textbf{Tra}ining with \textbf{T}ask \textbf{A}ugmentation, an approach that builds on two key ideas for effective leverage of unlabeled data. 
First, \strata uses \emph{task augmentation}, a novel technique that synthesizes a large amount of data for auxiliary-task fine-tuning from target-task unlabeled texts. Second, \strata performs \emph{self-training} by further fine-tuning the strong base model created by task augmentation on a broad distribution of pseudo-labeled data. Our experiments demonstrate that \strata can substantially improve sample efficiency across 12 few-shot benchmarks. Remarkably, on the \ssc{SST-2} sentiment dataset, \strata, with only 8 training examples per class, achieves comparable results to standard fine-tuning with 67K training examples. Our analyses reveal that task augmentation and self-training are both complementary and independently effective.

\blankfootnote{$\bigstar$ Work done as a student researcher at Google Brain.}
\end{abstract}
\section{Introduction}
\label{section:introduction}

\begin{figure}[t]
\centering
\includegraphics[width=0.48\textwidth]{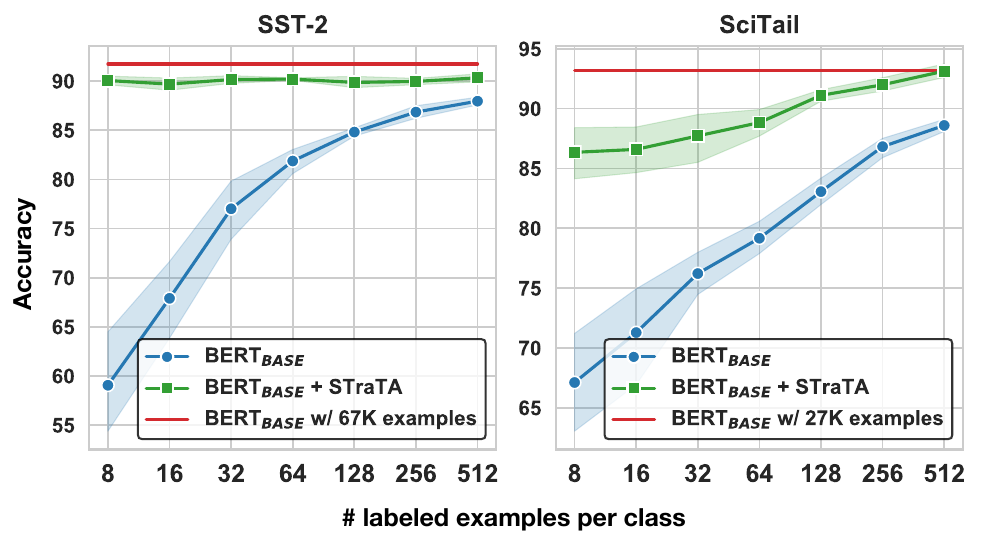}
\caption{Our \textbf{S}elf-\textbf{Tra}ining with \textbf{T}ask \textbf{A}ugmentation (\strata) approach substantially improves sample efficiency across different tasks. For example, when given only 8 labeled examples per class from the \ssc{SST-2} sentiment dataset, \strata is competitive with standard fine-tuning on 67K examples; on the \ssc{SciTail} entailment dataset, with 512 labeled examples per class, \ssc{STraTA} surpasses standard fine-tuning on 27K examples.
\label{fig:sample_efficiency}}
\vspace*{-2mm}
\end{figure}
Recent advances in \ssc{NLP} demonstrate the effectiveness of applying large-scale pretrained language models to downstream tasks~\cite{JDevlin19,YLiu19,ZYang19,ZLan20,CRaffel20,TBrown20,PHe21}. While these models have achieved state-of-the-art results on many \ssc{NLP} benchmarks, they struggle when given limited training data. For instance,~\citet{JDevlin19} find that \ssc{BERT} is prone to degenerate performance on small datasets. While enormous language models like \ssc{GPT-3}~\cite{TBrown20} exhibit the ability to solve a new task from only a few examples without any %
fine-tuning, their performance still lags far behind state-of-the-art fine-tuning results. %
Manually annotating large amounts of training data will likely improve performance but can also be prohibitively expensive to obtain for many tasks and domains. %
\begin{figure*}[t!]
\centering
\includegraphics[width=\textwidth]{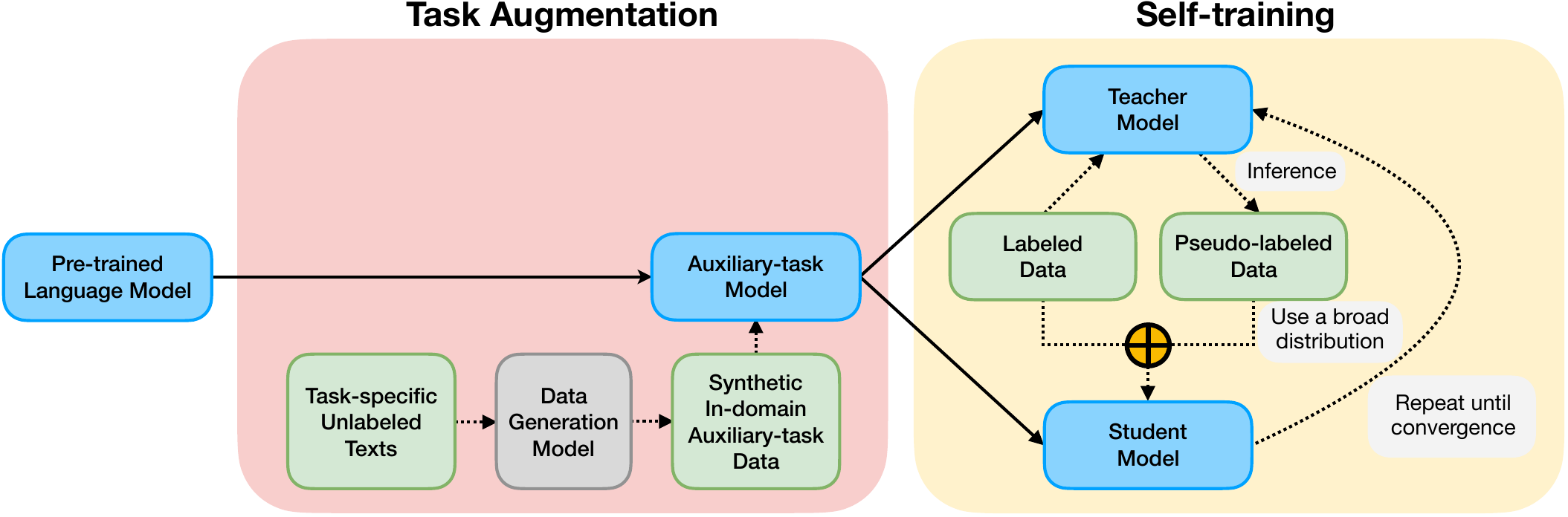}

\caption{An illustration of our \textbf{S}elf-\textbf{Tra}ining with \textbf{T}ask \textbf{A}ugmentation (\strata) approach. In task augmentation, we train an \ssc{NLI} data generation model and use it to synthesize a large amount of \emph{in-domain} \ssc{NLI} training data for each given target task, which is then used for auxiliary (intermediate) fine-tuning. Our self-training algorithm iteratively learns a better model using a concatenation of labeled and pseudo-labeled examples. At each iteration, we always start with the auxiliary-task model produced by task augmentation and train on a broad distribution of pseudo-labeled data.}

\vspace{-2mm}
\label{fig:ta_st}
\end{figure*}

In this paper, we propose \strata, an approach that combines two \emph{complementary} methods, \textbf{S}elf-\textbf{Tra}ining and \textbf{T}ask \textbf{A}ugmentation, 
to effectively leverage
unlabeled data, which is comparatively cheaper to obtain.\footnote{Code and pre-trained models available at\\ \href{https://github.com/google-research/google-research/tree/master/STraTA}{\ttfamily\myfontsize{https://github.com/google-research/\\google-research/tree/master/STraTA}}.}

At a high level, \emph{task augmentation} exploits unlabeled texts from the domain of a given target task to simulate a large amount of \textit{in-domain} training data for the \emph{auxiliary} task of  natural language inference (\ssc{NLI}), which is then used to train a given model before applying it to the target task. To achieve this, we first build an \ssc{NLI} data generator by fine-tuning a pre-trained generative language model on the \ssc{MNLI} dataset~\citep{AWilliams18} in a \textit{text-to-text} format. Then, given a target task (e.g., sentiment analysis) with unlabeled texts (e.g., \emph{his acting was really awful}), we use the \ssc{NLI} data generator to generate \ssc{NLI} examples (e.g., $[$\emph{his acting was really awful}, \emph{he gave an incredible performance}, \emph{contradiction}$]$). %
We show that task augmentation alone can significantly improve downstream performance across different tasks, generally outperforming other fine-tuning approaches, such as target-task language model fine-tuning~\cite{JHoward18,SGururangan20} and intermediate-task fine-tuning on \ssc{MNLI}~\cite{JPhang19}, in both high- and low-data regimes.

Having obtained a strong auxiliary-task model with task augmentation, \strata uses this model as a base model for \emph{self-training}. Specifically, at each at iteration, the base model is fine-tuned using the available labeled data for the target task. Then, the resulting model's predictions on unlabeled examples\footnote{We use the term \textit{unlabeled texts} to refer to pieces of text (e.g., sentences) from the target domain, and the term \textit{unlabeled examples} to refer to examples that can be annotated using the set of class labels for the target task.} are used as pseudo-labels to augment the original labeled data set. The newly formed labeled data set is then used to learn a better model in the next iteration, and this procedure is repeated for a number of iterations until a stopping criterion is reached. While self-training has been extensively studied~\cite{CRosenberg05,DMcClosky06,JHe20,QXie20b,JDu20}, our experiments reveal that using a strong base model and training on a broad distribution of pseudo-labeled data are key factors for successful deployment in \ssc{NLP}.

Using our \strata approach, we are able to significantly improve sample efficiency, in terms of both performance and variance, across 12 \ssc{NLP} benchmark datasets. For instance, on the \ssc{SST-2} sentiment dataset~\cite{RSocher13}, with only 8 training examples per class, we achieve comparable results to standard fine-tuning with 67K training examples (see Figure~\ref{fig:sample_efficiency}).%

Our main contributions are as follows:
\begin{enumerate}
    \item We propose task augmentation, a novel data augmentation-based fine-tuning method, and show its effectiveness in comparison to other competing fine-tuning approaches.
    \item We propose a simple yet effective self-training algorithm and highlight important ingredients for successful self-training, which we hope will enable the wider adoption of self-training in \ssc{NLP}.
    \item With \strata, we demonstrate the effectiveness of combining task augmentation and self-training in improving sample efficiency across \ssc{NLP} benchmarks.
\end{enumerate}

\section{Task augmentation}
Labeled data is often expensive and time-consuming to obtain, which motivates approaches that learn from both labeled and unlabeled data. %
More formally, assume we are given a target task $\bmmc{T}$ with a labeled data set $\bmmc{L}_{\bmmc{T}} = \{(\bm{x}_i, \bm{y}_i)\}_{i=1}^{M}$ and an unlabeled \text{data set } $\bmmc{U}_{\bmmc{T}} = \{(\bm{x}_j)\}_{j=1}^{N}$. The unlabeled \text{data } $\bmmc{U}_{\bmmc{T}}$ can be created artificially by removing the ground-truth labels $\bm{y}$ from $\bmmc{L}_{\bmmc{T}}$ (as in our main experiments), or it can come from additional unlabeled texts from the target domain or from related datasets/domains (see Section~\ref{sec:ood_unlabeled_data}).
Our methods, task augmentation and self-training, take advantage of the unlabeled \text{data } $\bmmc{U}_{\bmmc{T}}$ to maximize performance on the target task $\bmmc{T}$, even when the number of labeled examples $M$ is small (e.g., $M=16$). In this section, we first present a framework and implementation for task augmentation, which uses natural language inference (\ssc{NLI}) as an auxiliary (intermediate) training task to improve downstream performance.
\subsection{A framework for task augmentation}
Task augmentation builds on a recent body of \ssc{NLP} research on intermediate-task training~\citep{JPhang19,TVu20}, in which a pre-trained language model, such as \ssc{BERT}, is fine-tuned on an auxiliary task before the target task.\footnote{This process differs from traditional data augmentation approaches (e.g., lexical substitution, or back-translation), which yield negligible improvements when combined with large-scale pre-trained language models~\cite{JWei19,YYang20}.} In  previous work on intermediate fine-tuning, the auxiliary dataset used is a fixed target task-independent dataset, such as \ssc{MNLI} or \ssc{SQuAD}~\cite{PRajpurkar16}.
An obvious limitation of this choice is the domain mismatch between the auxiliary and target tasks, which our proposed task augmentation method addresses. %
More specifically, we fine-tune a pre-trained generative language model %
and use it to synthesize a large amount of \textit{in-domain} training data from $\bmmc{U}_{\bmmc{T}}$ for an auxiliary task $\bmmc{A}$, which is then used to improve performance of a model on the target task $\bmmc{T}$ (Figure~\ref{fig:ta_st}, left).\footnote{Traditional data augmentation is a special case of our framework where the auxiliary task is identical to the target task ($\bmmc{A} \equiv \bmmc{T}$).} In this work, we choose \ssc{NLI} as the auxiliary task for two main reasons: (1) \ssc{NLI} has been shown to be an effective auxiliary task for a variety of target tasks~\cite{AConneau17,JPhang19},
and (2) existing \ssc{NLI} datasets contain large training sets, which allows us to train a reliable data generator. 
\vspace{-1.5mm}
\paragraph{Generating synthetic \bssc{NLI} data:} To obtain an \ssc{NLI} data generator, we fine-tune the pre-trained \ssc{T5-3B} model~\cite{CRaffel20} %
on \ssc{MNLI}, which contains 393K sentence pairs labeled as \{\emph{entailment, contradiction, neutral}\}. %
We cast each \ssc{MNLI} training example $(\bm{sent}_A, \bm{sent}_B) \rightarrow \bm{label}$ into a \emph{text-to-text} format $(\bm{label, } \bm{sent}_A) \rightarrow \bm{sent}_B$ to obtain fine-tuning examples that look like $[$\emph{entailment, the facts are accessible to you $\rightarrow$ you have access to the facts}$]$.\footnote{We fine-tune a separate \fssc{T5} model per class label. To overcome biases in \fssc{MNLI} where the hypotheses are usually shorter than the premises, we also include reversed examples: $(\bm{reversed\;label, } \bm{sent}_B) \rightarrow \bm{sent}_A.$} We fine-tune \ssc{T5} on this dataset with a constant learning rate of 0.001 for $2^{16} = 65,536$ steps using the Adafactor optimizer~\cite{NShazeer18}.
The fine-tuned \ssc{T5} data generator produces augmented examples for all target datasets. Specifically, at inference time, we  feed the model an \ssc{NLI} label (e.g., \textit{entailment}) and an unlabeled sentence $\bm{x}_j$ from the target domain to produce some output sentence $\bm{x}_k$: $(entailment, \bm{x}_j) \rightarrow \bm{x}_k$ (see Appendix~\ref{appendix:b} for example outputs). Data for intermediate fine-tuning is then formed by creating examples like $(\bm{x}_j, \bm{x}_k) \rightarrow entailment$. This approach has several advantages: (1) training labels are \textit{free}, and (2) by using overgeneration, we can produce a large amount of \textit{in-domain} \ssc{NLI} training data even for target tasks with small datasets.
\vspace{-1.5mm}
\paragraph{Overgeneration and filtering:} Following~\citet{RPuri20}, we perform overgeneration and filtering to increase the quantity and quality of the synthetic \ssc{NLI} training data. Concretely, we generate 100 output samples per input with top-$k$ ($k = 40$) sampling (duplicates are removed) and use a \ssc{BERT} model fine-tuned on \ssc{MNLI} (in the original format) as an \ssc{NLI} classifier to filter synthetic training examples. We keep a synthetic example if the \ssc{NLI} classifier produces the same label as that fed to the \ssc{NLI} data generator and is also confident about its prediction.\footnote{We use an example when its predicted probability exceeding a certain threshold $\tau$. We choose a value for $\tau$ in $[0.3, 0.4, \ldots,0.9]$ for each target task based on performance on the original \fssc{MNLI} development set.} For all experiments, we perform intermediate fine-tuning on examples from both the original \ssc{MNLI} dataset and the final filtered task augmentation dataset.\footnote{A two-stage intermediate fine-tuning procedure where the model is first trained on the synthetic data before being fine-tuned on the original data typically works better, and this is used in our experiments.}

\section{Self-training}
While task augmentation uses unlabeled texts to produce synthetic data for an intermediate task, self-training is a \emph{complementary} approach that improves a model by training directly on the target task using pseudo-labeled examples. In this section, we 
explore a simple self-training algorithm in which a model learns to improve itself using its predictions on unlabeled examples from a given target task. Our method differs from traditional self-training methods in that we leverage a strong base model and allow it to learn from all available pseudo-labeled examples at every iteration, regardless of model confidence.
Formally, given a target task $\bmmc{T}$ with a small labeled data set $\bmmc{L} = \{(\bm{x}_i, \bm{y}_i)\}_{i=1}^{M}$ and an unlabeled data \text{set } $\bmmc{U} = \{(\bm{x}_j)\}_{j=1}^{N}$, where $M \ll N$, we summarize our self-training algorithm in Algorithm~\ref{algo:selftraining}. 
\begin{algorithm}[t]
\SetAlgoLined
\renewcommand{\arraystretch}{1.5}
\DontPrintSemicolon
\Init{}{
\small
\begin{tabularx}{0.42\textwidth}{X}
$t = 0$ \\
Form a base model $\bm{f}_0$, which is initialized with pre-trained parameters from a pre-training/intermediate fine-tuning stage, and then learn a teacher model $\bm{f}_1$ by training $\bm{f}_0$ on the original labeled data set $\bmmc{L}$. 
\end{tabularx}
}
\Repeat{convergence or the maximum number of iterations is reached}{
\small
\begin{tabularx}{0.42\textwidth}{X}
$t=t+1$ \\
1. Use the current teacher model $\bm{f}_t$ to annotate (for $t=1$) or re-annotate (for $t > 1$) all of the examples \text{in } $\bmmc{U}$ to obtain a \text{set } $\bmmc{\widetilde{U}}$ of pseudo-labeled examples. \\
2. %
Add the whole \text{set } $\bmmc{\widetilde{U}}$ of pseudo-labeled examples to the original labeled data set $\bmmc{L}$ to form a new labeled data set.
\\
3. Learn a student model $\bm{f}_{t+1}$ by training the base model $\bm{f}_0$ on the current labeled data set and optionally fine-tune it on $\bmmc{L}$. The resulting student model $\bm{f}_{t+1}$ is used as a teacher for the next iteration. \\
\end{tabularx}
}
\caption{Our self-training algorithm}
\label{algo:selftraining}
\end{algorithm}
\renewcommand{\arraystretch}{1}

\vspace{-1.5mm}
\paragraph{Starting with a strong base model:} An important ingredient in self-training algorithms is the base model $\bm{f}_0$. Successful self-training typically requires a good base model, which can provide a large proportion of ``correct'' predictions or pseudo-labels on unlabeled examples; otherwise, errors can be propagated or magnified in later stages of self-training. %
At each self-training iteration, we always start from the same base model $\bm{f}_0$, which is initialized with pre-trained parameters from a pre-training/intermediate fine-tuning stage (e.g., the auxiliary task training stage in task augmentation),\footnote{We find empirically that starting from the base model $\bm{f}_0$ works better than from the model $\bm{f}_{t-1}$ obtained in the previous iteration.} and then fine-tune all of its parameters using the available labeled and pseudo-labeled data.\footnote{\citet{JHe20} find that further fine-tuning the resulting model on the original labeled data set $\bmmc{L}$ improves machine translation models. We use development set performance to decide whether or not to perform this fine-tuning step for each dataset.}%
\vspace{-1.5mm}
\paragraph{Self-training on a broad distribution of pseudo-labeled data:} 

Another important factor is the selection of pseudo-labeled examples at each self-training iteration. Traditional self-training approaches usually select a small set of examples where the current teacher model $\bm{f}_t$ is highly confident (e.g., the probability of the predicted class label is above a threshold) to add to the labeled data set at each iteration until the unlabeled data \text{pool } $\bmmc{U}$ is exhausted. This can be problematic as state-of-the-art language models like \ssc{BERT} are overconfident and poorly calibrated~\citep{ZJiang21}. In preliminary experiments, we tried several calibration methods, including temperature scaling~\cite{CGuo17}, label smoothing~\cite{RMuller19}, and confidence penalties~\cite{GPereyra17}, but all of which failed to fully address this problem. %
Instead, we encourage learning from a ``natural'' broad distribution of pseudo-labeled data by adding the whole \text{set } $\bmmc{\widetilde{U}}$ of pseudo-labeled examples to the original labeled data set $\bmmc{L}$ at each self-training iteration.\footnote{We find that removing examples with the lowest-confidence pseudo labels can be helpful for some tasks. One can use a development set, upon availability, to assess if this filtering is necessary.} At each iteration $t > 1$, we also re-annotate all of the examples in the original unlabeled data \text{pool } $\bmmc{U}$  with $\bm{f}_{t}$, as we expect $\bm{f}_{t}$ is better than $\bm{f}_{t-1}$.

\section{Experiments}
We perform experiments across 12 different \ssc{NLP} datasets and three different data regimes (including a few-shot setting). Task augmentation consistently improves over prior fine-tuning approaches in all three regimes, and the combination of self-training and task augmentation, \strata, results in higher performance and lower variance than competing approaches when given only 8 labeled examples per class from each dataset.
\subsection{Datasets \& data regimes}
\label{sec:data_regimes}
\begin{table}[t!]
\centering
\begin{adjustbox}{max width=0.40\textwidth}
\begin{tabular}{l r}
\toprule
\multicolumn{1}{l}{\textbf{Task}} & \multicolumn{1}{r}{\textbf{$|$Train$|$}} \\ [0.5ex]
\multicolumn{2}{l}{\emph{text classification/regression}} \\ [0.5ex] 
\ssc{SNLI}~\cite{SBowman15} & 570K \\
\ssc{MNLI}~\cite{AWilliams18} & 393K \\
\ssc{QQP}~\cite{SIyer17} & 364K \\
\ssc{QNLI}~\cite{AWang19} & 105K \\
\ssc{SST-2}~\cite{RSocher13} &  67K \\
\ssc{SciTail}~\cite{TKhot18} & 27K \\
\ssc{SST-5}~\cite{RSocher13} &  8.5K \\
\ssc{STS-B}~\cite{DCer17} & 7K \\
\ssc{SICK-E}~\cite{MMarelli14} & 4.5K \\
\ssc{SICK-R}~\cite{MMarelli14} & 4.5K \\
\ssc{CR}~\cite{MHu04} & 4K \\
\ssc{MRPC}~\cite{WDolan05} & 3.7K \\
\ssc{RTE}~\cite[][et seq.]{IDagan06} & 2.5K \\
\bottomrule
\end{tabular}
\end{adjustbox}
\caption{Datasets used in our experiments.%
}
\vspace*{-2mm}
\label{table:datasets}
\end{table}
The datasets used in our study (Table~\ref{table:datasets})\footnote{Appendix~\ref{appendix:a} contains more details about characteristics and associated evaluation metrics for each dataset.}  come from two common language understanding benchmarks: \ssc{GLUE}~\cite{AWang19} and \ssc{SentEval}~\cite{AConneau18}. Due to restricted test set access for \ssc{GLUE} datasets, we held out a small subset of the training set for validation and report results on the original development set for each task. The training set without ground-truth labels is used as unlabeled \text{data } $\bmmc{U}_{\bmmc{T}}$.

We consider three data regimes by varying the amount of labeled training data across the downstream tasks: \ssc{Full} (all labeled training data), \ssc{Limited} (1024 random labeled training examples), and \ssc{Few-shot} (8 random labeled training examples per class).\footnote{For regression tasks, we partition the output interval $[0,5]$ into five bins and sample 8 examples from each bin.} Since fine-tuning \ssc{BERT} can be unstable on small datasets~\cite{JDevlin19},  we perform 10 random restarts where there are less than 10K training examples and report the mean and standard deviation.\footnote{We resample examples for each restart.} Since large development sets are impractical in  low-resource settings~\cite{AOliver18,KKann19}, we randomly sample 256 development examples for each task in the \ssc{Limited} and \ssc{Few-shot} regimes. Additionally, in the \ssc{Few-shot} regime, we experiment with a real-world scenario where there is no development set access.
\subsection{Setup}
As in~\citet{JDevlin19}, our input format for all tasks contains a \textsc{[cls]} token followed by a single text segment or a concatenation of text segments (e.g., a premise-hypothesis pair) separated with a \textsc{[sep]} token. We feed the final \textsc{[cls]} representation into a task-specific classification layer and fine-tune all the parameters end-to-end on the downstream tasks. For both fine-tuning and self-training, we perform early stopping based on development set performance. %
We use the Transformers library~\cite{TWolf19} and its recommended hyperparameters for all experiments.\footnote{While individual task performance can likely be further improved with more involved hyperparameter tuning, %
we standardize hyperparameters across tasks to cut down on computational expense. Our experiments were conducted on Google Cloud with 100\% renewable energy.}

\subsection{Methods} 
We experiment with task augmentation (\bssc{TA}) and self-training (\bssc{ST}) individually, as well as the combined approach \ssc{STraTA}, which uses the auxiliary-task model from task augmentation as the base model for self-training. We compare our methods to the following baselines:
\vspace{-1.5mm}
\paragraph{\bssc{LMFT} \& \bssc{ITFT$_{\smallsup{MNLI}}$}:}
We compare our methods against commonly-used fine-tuning approaches, including target-task language model fine-tuning~\cite[\ssc{LMFT};][]{JHoward18,SGururangan20}---in which a model is first trained with the language model objective on task-specific unlabeled data before being fine-tuned on the target task---and intermediate-task fine-tuning on \ssc{MNLI}~\cite[\ssc{ITFT$_{\smallsup{MNLI}}$};][]{JPhang19}---which first trains a model on %
\ssc{MNLI} before fine-tuning it on the target task.
\begin{table*}[t]
\centering
\begin{adjustbox}{max width=0.99\textwidth}
\begin{tabular}{ l l l l l  l l l l  l l l l}
\toprule
\textbf{Model} & \sc{SNLI} & \sc{QQP} & \sc{QNLI} & \sc{SST-2} & \sc{SciTail} & \sc{SST-5} & \sc{STS-B} & \sc{SICK-E} & \sc{SICK-R} & \sc{CR} & \sc{MRPC} & \sc{RTE} \\
\midrule
\midrule
\multicolumn{13}{c}{\sc{Full}} \\
\lbertlarge\ & \textcolor{black}{91.1} & \textcolor{black}{88.4} & 
\textcolor{black}{91.9} & 
\textcolor{black}{92.4} & 
\textcolor{black}{95.3} & \textcolor{black}{53.7$_{\smallsup{0.9}}$} & \textcolor{black}{89.6$_{\smallsup{0.2}}$} & \textcolor{black}{87.9$_{\smallsup{0.6}}$} & \textcolor{black}{84.4$_{\smallsup{0.4}}$} & \textcolor{black}{91.7$_{\smallsup{0.6}}$} & \textcolor{black}{89.0$_{\smallsup{0.8}}$} & \textcolor{black}{68.6$_{\smallsup{7.2}}$} \\
\hspace{10pt}$\text{+ \sc{LMFT}}$ & \textcolor{black}{91.0} &
\textcolor{black}{88.1} & \textcolor{black}{90.4} & \textcolor{black}{93.5} & \textcolor{black}{95.3} & \textcolor{black}{54.0$_{\smallsup{0.4}}$} & \textcolor{black}{89.5$_{\smallsup{0.2}}$} & \textcolor{black}{87.7$_{\smallsup{0.5}}$} & \textcolor{black}{84.0$_{\smallsup{0.5}}$} & \textcolor{black}{91.6$_{\smallsup{0.8}}$} & \textcolor{black}{89.5$_{\smallsup{1.0}}$} & \textcolor{black}{66.5$_{\smallsup{7.3}}$} \\
\hspace{10pt}\text{+ \sc{ITFT$_{\textsc{MNLI}}$}} &
\textcolor{black}{91.1} & \textcolor{black}{88.2} & \textcolor{black}{91.6} & \textcolor{black}{93.5} & \textcolor{black}{96.5} & \textcolor{black}{54.0$_{\smallsup{0.8}}$} & \textcolor{black}{90.3$_{\smallsup{0.3}}$} & \textcolor{black}{89.9$_{\smallsup{0.2}}$} & \textcolor{black}{86.3$_{\smallsup{0.3}}$} & \textcolor{black}{92.0$_{\smallsup{0.6}}$} & \textcolor{black}{89.7$_{\smallsup{0.9}}$} & \textcolor{black}{82.3$_{\smallsup{1.4}}$} \\
\hspace{10pt}\text{+ \sc{TA}} & \textcolor{black}{\textbf{91.9}} & \textcolor{black}{\textbf{88.5}} & \textcolor{black}{\textbf{92.5}} & \textcolor{black}{\textbf{94.7}} & \textcolor{black}{\textbf{96.9}} & \textcolor{black}{\textbf{55.7}$_{\smallsup{0.8}}$} & \textcolor{black}{\textbf{90.9}$_{\smallsup{0.2}}$} & \textcolor{black}{\textbf{90.7}$_{\smallsup{0.3}}$} & \textcolor{black}{\textbf{87.0}$_{\smallsup{0.3}}$} & \textcolor{black}{\textbf{93.3}$_{\smallsup{0.6}}$} \textcolor{black} & {\textbf{90.8}$_{\smallsup{0.7}}$} & \textcolor{black}{\textbf{83.8}$_{\smallsup{1.1}}$}\\
\midrule
\midrule
\multicolumn{13}{c}{\sc{Limited} \emph{(1024 total training examples)}}\\
\lbertlarge\ & \textcolor{black}{77.4$_{\smallsup{0.6}}$} & \textcolor{black}{74.1$_{\smallsup{1.0}}$} & \textcolor{black}{81.7$_{\smallsup{0.9}}$} & \textcolor{black}{89.8$_{\smallsup{0.6}}$} & \textcolor{black}{90.9$_{\smallsup{0.7}}$} & \textcolor{black}{49.1$_{\smallsup{1.3}}$} & \textcolor{black}{88.2$_{\smallsup{0.4}}$} & \textcolor{black}{84.8$_{\smallsup{0.7}}$} & \textcolor{black}{80.2$_{\smallsup{0.4}}$} & \textcolor{black}{91.2$_{\smallsup{0.6}}$} & \textcolor{black}{85.7$_{\smallsup{1.7}}$} & \textcolor{black}{66.8$_{\smallsup{2.7}}$} \\
\hspace{10pt}$\text{+ \sc{LMFT}}$ &
\textcolor{black}{75.8$_{\smallsup{1.5}}$} & \textcolor{black}{71.6$_{\smallsup{0.5}}$} & \textcolor{black}{80.5$_{\smallsup{2.0}}$} & \textcolor{black}{88.9$_{\smallsup{0.8}}$} & \textcolor{black}{87.7$_{\smallsup{2.3}}$} & \textcolor{black}{49.2$_{\smallsup{3.1}}$} & \textcolor{black}{88.4$_{\smallsup{0.4}}$} & \textcolor{black}{83.2$_{\smallsup{0.6}}$} & \textcolor{black}{78.5$_{\smallsup{0.6}}$} & \textcolor{black}{90.9$_{\smallsup{0.7}}$} & \textcolor{black}{84.9$_{\smallsup{1.1}}$} & \textcolor{black}{65.2$_{\smallsup{3.4}}$} \\
\hspace{10pt}\text{+ \sc{ITFT$_{\textsc{MNLI}}$}} &
\textcolor{black}{85.2$_{\smallsup{0.4}}$} & \textcolor{black}{74.0$_{\smallsup{0.5}}$} & \textcolor{black}{83.5$_{\smallsup{0.5}}$} & \textcolor{black}{90.0$_{\smallsup{0.8}}$} & \textcolor{black}{92.1$_{\smallsup{1.1}}$} & \textcolor{black}{49.4$_{\smallsup{1.2}}$} & \textcolor{black}{87.8$_{\smallsup{0.8}}$} & \textcolor{black}{88.8$_{\smallsup{0.5}}$} & \textcolor{black}{83.2$_{\smallsup{0.7}}$} & \textcolor{black}{91.3$_{\smallsup{0.7}}$} & \textcolor{black}{86.4$_{\smallsup{0.9}}$} & \textcolor{black}{81.1$_{\smallsup{1.3}}$} \\
\hspace{10pt}\text{+ \sc{TA}} & \textcolor{black}{\textbf{87.3}$_{\smallsup{0.3}}$} & \textcolor{black}{\textbf{75.7}$_{\smallsup{0.5}}$} & \textcolor{black}{\textbf{85.0}$_{\smallsup{0.5}}$} & \textcolor{black}{\textbf{91.7}$_{\smallsup{0.7}}$} & \textcolor{black}{\textbf{92.3}$_{\smallsup{1.1}}$} & \textcolor{black}{\textbf{51.4}$_{\smallsup{1.0}}$} & \textcolor{black}{\textbf{89.0}$_{\smallsup{0.6}}$} & \textcolor{black}{\textbf{89.4}$_{\smallsup{0.4}}$} & \textcolor{black}{\textbf{84.3}$_{\smallsup{0.4}}$} & \textcolor{black}{\textbf{92.6}$_{\smallsup{0.6}}$} & \textcolor{black}{\textbf{88.0}$_{\smallsup{0.8}}$} & \textcolor{black}{\textbf{82.9}$_{\smallsup{1.8}}$} \\
\midrule
\midrule
\multicolumn{13}{c}{\sc{Few-Shot} \emph{(8 training examples per class)}}\\
\lbertbase\ &
 \textcolor{black}{43.7$_{\smallsup{2.2}}$} & \textcolor{black}{55.9$_{\smallsup{6.5}}$} & \textcolor{black}{59.0$_{\smallsup{10.9}}$} & \textcolor{black}{59.1$_{\smallsup{8.4}}$} & \textcolor{black}{67.1$_{\smallsup{6.6}}$} & \textcolor{black}{30.5$_{\smallsup{2.0}}$} & \textcolor{black}{73.6$_{\smallsup{4.5}}$} & \textcolor{black}{61.3$_{\smallsup{4.1}}$} & \textcolor{black}{59.7$_{\smallsup{2.7}}$} & \textcolor{black}{65.2$_{\smallsup{8.2}}$} & \textcolor{black}{72.4$_{\smallsup{10.2}}$} & \textcolor{black}{51.4$_{\smallsup{2.5}}$} \\
\hspace{10pt}$\text{+ \sc{LMFT}}$ &
\textcolor{black}{45.2$_{\smallsup{3.9}}$} & \textcolor{black}{57.2$_{\smallsup{6.2}}$} & \textcolor{black}{57.6$_{\smallsup{9.1}}$} & \textcolor{black}{64.9$_{\smallsup{8.7}}$} & \textcolor{black}{64.0$_{\smallsup{8.0}}$} & \textcolor{black}{33.4$_{\smallsup{1.9}}$} & \textcolor{black}{75.4$_{\smallsup{4.4}}$} & \textcolor{black}{59.3$_{\smallsup{4.0}}$} & \textcolor{black}{58.3$_{\smallsup{2.0}}$} & \textcolor{black}{72.4$_{\smallsup{6.0}}$} & \textcolor{black}{73.9$_{\smallsup{8.6}}$} & \textcolor{black}{50.9$_{\smallsup{3.9}}$} \\
\hspace{10pt}\text{+ \sc{ITFT$_{\textsc{MNLI}}$}} &
\textcolor{black}{75.2$_{\smallsup{5.7}}$} & \textcolor{black}{63.7$_{\smallsup{7.0}}$} & \textcolor{black}{62.8$_{\smallsup{5.1}}$} & \textcolor{black}{76.8$_{\smallsup{7.2}}$} & \textcolor{black}{75.8$_{\smallsup{5.6}}$} & \textcolor{black}{35.0$_{\smallsup{2.6}}$} & \textcolor{black}{80.2$_{\smallsup{1.1}}$} & \textcolor{black}{80.4$_{\smallsup{1.9}}$} & \textcolor{black}{73.5$_{\smallsup{2.7}}$} & \textcolor{black}{79.2$_{\smallsup{3.6}}$} & \textcolor{black}{74.3$_{\smallsup{8.0}}$} & \textcolor{black}{62.2$_{\smallsup{13.5}}$} \\
\hspace{10pt}\text{+ \sc{TA}} & \textcolor{black}{83.3$_{\smallsup{0.8}}$} & \textcolor{black}{68.7$_{\smallsup{1.5}}$} & \textcolor{black}{70.1$_{\smallsup{3.4}}$} & \textcolor{black}{80.3$_{\smallsup{6.6}}$} & \textcolor{black}{78.5$_{\smallsup{3.2}}$} & \textcolor{black}{37.4$_{\smallsup{3.0}}$} & \textcolor{black}{80.7$_{\smallsup{1.5}}$} & \textcolor{black}{81.1$_{\smallsup{2.4}}$} & \textcolor{black}{75.9$_{\smallsup{1.8}}$} & \textcolor{black}{86.5$_{\smallsup{2.2}}$} & \textcolor{black}{74.5$_{\smallsup{6.5}}$} & \textcolor{black}{67.6$_{\smallsup{7.1}}$} \\
\hspace{10pt}\text{+ \sc{ST}} &
\textcolor{black}{65.0$_{\smallsup{5.8}}$} & \textcolor{black}{69.9$_{\smallsup{5.9}}$} &
\textcolor{black}{71.6$_{\smallsup{11.3}}$} &
\textcolor{black}{62.7$_{\smallsup{10.4}}$} & \textcolor{black}{68.6$_{\smallsup{8.3}}$} & \textcolor{black}{33.9$_{\smallsup{3.5}}$} &
\textcolor{black}{80.5$_{\smallsup{2.2}}$} &
\textcolor{black}{68.1$_{\smallsup{4.5}}$} &
\textcolor{black}{64.0$_{\smallsup{2.4}}$} &
\textcolor{black}{78.2$_{\smallsup{6.3}}$} &
\textcolor{black}{80.5$_{\smallsup{1.8}}$} &
\textcolor{black}{50.7$_{\smallsup{3.1}}$} \\
\hspace{10pt}\text{+ \sc{ITFT$_{\textsc{MNLI}}$ + ST}} &
\textcolor{black}{83.2$_{\smallsup{0.3}}$} &
\textcolor{black}{70.7$_{\smallsup{5.9}}$} &
\textcolor{black}{81.5$_{\smallsup{1.2}}$} &
\textcolor{black}{88.0$_{\smallsup{2.1}}$} & \textcolor{black}{83.7$_{\smallsup{4.4}}$} & \textcolor{black}{39.5$_{\smallsup{2.0}}$} &
\textcolor{black}{84.2$_{\smallsup{0.8}}$} &
\textcolor{black}{81.8$_{\smallsup{2.6}}$} & \textcolor{black}{75.8$_{\smallsup{2.2}}$} & \textcolor{black}{85.6$_{\smallsup{2.3}}$} &
\textcolor{black}{80.6$_{\smallsup{1.2}}$} &
\textcolor{black}{62.5$_{\smallsup{12.0}}$} \\
\hspace{10pt}\text{+ \lstrata} &
\textcolor{black}{\textbf{85.7}$_{\smallsup{0.2}}$} & \textcolor{black}{\textbf{74.5}$_{\smallsup{0.4}}$} &
\textcolor{black}{\textbf{82.1}$_{\smallsup{0.5}}$} &
\textcolor{black}{\textbf{90.1}$_{\smallsup{0.8}}$} & \textcolor{black}{\textbf{86.3}$_{\smallsup{3.5}}$} & \textcolor{black}{\textbf{41.3}$_{\smallsup{1.5}}$} &
\textcolor{black}{\textbf{84.7}$_{\smallsup{0.5}}$} &
\textcolor{black}{\textbf{84.9}$_{\smallsup{1.2}}$} & \textcolor{black}{\textbf{77.6}$_{\smallsup{1.6}}$} &
\textcolor{black}{\textbf{90.5}$_{\smallsup{0.8}}$} &
\textcolor{black}{\textbf{81.0}$_{\smallsup{0.8}}$} &
\textcolor{black}{\textbf{70.6}$_{\smallsup{2.4}}$} \\
\midrule
\lbertlarge\ &
\textcolor{black}{43.1$_{\smallsup{4.4}}$} & \textcolor{black}{58.5$_{\smallsup{4.7}}$} & \textcolor{black}{64.4$_{\smallsup{6.1}}$} & \textcolor{black}{66.1$_{\smallsup{8.7}}$} & \textcolor{black}{68.8$_{\smallsup{9.5}}$} & \textcolor{black}{35.2$_{\smallsup{1.3}}$} & \textcolor{black}{74.6$_{\smallsup{3.8}}$} & \textcolor{black}{66.5$_{\smallsup{4.5}}$} & \textcolor{black}{66.6$_{\smallsup{3.3}}$} & \textcolor{black}{72.0$_{\smallsup{6.0}}$} & \textcolor{black}{79.9$_{\smallsup{2.0}}$} & \textcolor{black}{53.1$_{\smallsup{3.3}}$} \\
\hspace{10pt}$\text{+ \sc{LMFT}}$ &
\textcolor{black}{39.6$_{\smallsup{2.6}}$} & \textcolor{black}{52.7$_{\smallsup{4.7}}$} & \textcolor{black}{52.2$_{\smallsup{1.6}}$} & \textcolor{black}{66.3$_{\smallsup{9.3}}$} & \textcolor{black}{66.4$_{\smallsup{10.6}}$} & \textcolor{black}{36.8$_{\smallsup{2.9}}$} & \textcolor{black}{75.4$_{\smallsup{9.4}}$} & \textcolor{black}{58.8$_{\smallsup{6.9}}$} & \textcolor{black}{51.6$_{\smallsup{7.0}}$} & \textcolor{black}{75.6$_{\smallsup{5.9}}$} & \textcolor{black}{80.5$_{\smallsup{2.4}}$} & \textcolor{black}{52.8$_{\smallsup{4.8}}$} \\
\hspace{10pt}\text{+ \sc{ITFT$_{\textsc{MNLI}}$}} & \textcolor{black}{79.9$_{\smallsup{3.1}}$} & \textcolor{black}{62.6$_{\smallsup{9.0}}$} & \textcolor{black}{64.5$_{\smallsup{4.4}}$} & \textcolor{black}{80.7$_{\smallsup{5.0}}$} & \textcolor{black}{72.3$_{\smallsup{11.2}}$} & \textcolor{black}{36.4$_{\smallsup{2.1}}$} & \textcolor{black}{75.5$_{\smallsup{4.0}}$} & \textcolor{black}{77.8$_{\smallsup{3.8}}$} & \textcolor{black}{73.5$_{\smallsup{2.8}}$} & \textcolor{black}{82.6$_{\smallsup{3.0}}$} & \textcolor{black}{72.8$_{\smallsup{7.9}}$} & \textcolor{black}{69.7$_{\smallsup{14.6}}$} \\
\hspace{10pt}\text{+ \sc{TA}} & \textcolor{black}{84.8$_{\smallsup{0.7}}$} & \textcolor{black}{64.6$_{\smallsup{6.3}}$} & \textcolor{black}{71.5$_{\smallsup{4.0}}$} & \textcolor{black}{85.5$_{\smallsup{1.4}}$} & \textcolor{black}{79.0$_{\smallsup{4.5}}$} & \textcolor{black}{38.5$_{\smallsup{3.0}}$} & \textcolor{black}{78.9$_{\smallsup{2.4}}$} & \textcolor{black}{81.2$_{\smallsup{3.9}}$} & \textcolor{black}{77.5$_{\smallsup{1.4}}$} & \textcolor{black}{88.6$_{\smallsup{1.3}}$} & \textcolor{black}{78.2$_{\smallsup{6.6}}$} & \textcolor{black}{77.0$_{\smallsup{6.3}}$} \\
\hspace{10pt}\text{+ \sc{ST}}
& \textcolor{black}{69.3$_{\smallsup{9.2}}$} &
\textcolor{black}{74.3$_{\smallsup{1.2}}$} &
\textcolor{black}{85.4$_{\smallsup{1.7}}$} &
\textcolor{black}{81.9$_{\smallsup{12.2}}$} &
\textcolor{black}{79.9$_{\smallsup{4.8}}$} &
\textcolor{black}{42.0$_{\smallsup{1.5}}$} &
\textcolor{black}{82.8$_{\smallsup{2.3}}$} &
\textcolor{black}{77.3$_{\smallsup{3.1}}$} & \textcolor{black}{73.1$_{\smallsup{2.3}}$} &
\textcolor{black}{88.1$_{\smallsup{1.3}}$} &
\textcolor{black}{81.2$_{\smallsup{0.5}}$} &
\textcolor{black}{53.9$_{\smallsup{4.3}}$} \\
\hspace{10pt}\text{+ \sc{ITFT$_{\textsc{MNLI}}$ + ST}} & \textcolor{black}{85.4$_{\smallsup{0.3}}$} &
\textcolor{black}{74.8$_{\smallsup{0.7}}$} &
\textcolor{black}{86.1$_{\smallsup{1.1}}$} &
\textcolor{black}{89.7$_{\smallsup{0.7}}$} &
\textcolor{black}{86.2$_{\smallsup{4.2}}$} &
\textcolor{black}{42.2$_{\smallsup{2.0}}$} &
\textcolor{black}{84.1$_{\smallsup{1.7}}$} &
\textcolor{black}{84.3$_{\smallsup{2.0}}$} & \textcolor{black}{78.4$_{\smallsup{1.3}}$} &
\textcolor{black}{89.3$_{\smallsup{1.0}}$} &
\textcolor{black}{81.4$_{\smallsup{1.2}}$} &
\textcolor{black}{72.7$_{\smallsup{5.4}}$} \\
\hspace{10pt}\text{+ \lstrata} &
\textcolor{black}{\textbf{87.3}$_{\smallsup{0.3}}$} &
\textcolor{black}{\textbf{75.1}$_{\smallsup{0.2}}$} &
\textcolor{black}{\textbf{86.4}$_{\smallsup{0.8}}$} &
\textcolor{black}{\textbf{91.7}$_{\smallsup{0.7}}$} &
\textcolor{black}{\textbf{87.3}$_{\smallsup{2.9}}$} &
\textcolor{black}{\textbf{43.0}$_{\smallsup{2.3}}$} &
\textcolor{black}{\textbf{84.5}$_{\smallsup{1.6}}$} &
\textcolor{black}{\textbf{86.3}$_{\smallsup{1.8}}$} & \textcolor{black}{\textbf{79.0}$_{\smallsup{1.0}}$} &
\textcolor{black}{\textbf{90.0}$_{\smallsup{0.6}}$} &
\textcolor{black}{\textbf{81.5}$_{\smallsup{0.7}}$} &
\textcolor{black}{\textbf{77.1}$_{\smallsup{5.4}}$} \\
\midrule
\multicolumn{13}{c}{\emph{Prompt-based~\cite[LM-BFF;][]{TGao20} and entailment-based~\cite[EFL;][]{SWang21} fine-tuning approaches}} \\
\lrobertalarge\ & \textcolor{black}{38.4$_{\smallsup{1.3}}$} &
\textcolor{black}{58.8$_{\smallsup{9.9}}$} &
\textcolor{black}{52.7$_{\smallsup{1.8}}$} &
\textcolor{black}{60.5$_{\smallsup{3.1}}$} & 
-- & 
-- &
\textcolor{black}{24.5$_{\smallsup{8.4}}$} & 
-- & 
-- &
\textcolor{black}{61.9$_{\smallsup{5.1}}$} &
\textcolor{black}{76.1$_{\smallsup{3.9}}$} &
\textcolor{black}{55.0$_{\smallsup{1.3}}$} \\

\hspace{10pt}\text{+ \sc{LM-BFF}} & \textcolor{black}{52.0$_{\smallsup{1.7}}$} &
\textcolor{black}{\textbf{68.2}$_{\smallsup{1.2}}$} &
\textcolor{black}{61.8$_{\smallsup{3.2}}$} &
\textcolor{black}{79.9$_{\smallsup{6.0}}$} & 
-- & 
-- &
\textcolor{black}{66.0$_{\smallsup{3.2}}$} & 
-- & 
-- &
\textcolor{black}{88.6$_{\smallsup{2.3}}$} &
\textcolor{black}{\textbf{78.5}$_{\smallsup{2.3}}$} &
\textcolor{black}{63.3$_{\smallsup{2.1}}$} \\

\hspace{10pt}\text{+ \sc{EFL}} &
\textcolor{black}{\textbf{81.0}$_{\smallsup{1.1}}$} &
\textcolor{black}{67.3$_{\smallsup{2.6}}$} &
\textcolor{black}{\textbf{68.0}$_{\smallsup{3.4}}$} &
\textcolor{black}{\textbf{90.8}$_{\smallsup{1.0}}$} & 
-- & 
-- &
\textcolor{black}{\textbf{71.0}$_{\smallsup{1.3}}$} & 
-- & 
-- &
\textcolor{black}{\textbf{92.3}$_{\smallsup{0.4}}$} &
\textcolor{black}{76.2$_{\smallsup{1.3}}$} &
\textcolor{black}{\textbf{85.8}$_{\smallsup{0.9}}$} \\
\bottomrule
\end{tabular}
\end{adjustbox}
\caption{\strata significantly improves results across 12 \ssc{NLP} benchmark datasets (numbers in the subscript  indicate the standard deviation across 10 random seeds). See Appendix~\ref{appendix:d} for full results.}
\label{tbl:main_results}
\vspace{-2mm}
\end{table*}

\vspace{-1.5mm}
\paragraph{\bssc{LM-BFF} \& \bssc{EFL}:}
We also include results from recent work on prompt-based~\cite[\ssc{LM-BFF};][]{TGao20} and entailment-based~\cite[\ssc{EFL};][]{SWang21} fine-tuning,\footnote{Results taken from~\citet{SWang21}.} which has been shown to outperform the \ssc{GPT-3}-style ``in-context learning'' approach~\cite{TBrown20} for few-shot learning. These approaches do not assume access to task-specific unlabeled data and are not directly comparable to our methods due to differences in model architecture and experimental settings.
\vspace{-1.5mm}
\paragraph{\bssc{SentAug-ST}:}
Closely related to our work, \citet{JDu20} propose a sentence augmentation method that retrieves a large amount of ``in-domain'' data for a given task from a large bank of Web sentences. A base model trained on task-specific labeled data is applied to obtain pseudo-labels for the retrieved sentences, which are then added to the original labeled set to train a better model.

\subsection{Results and Discussion}
Table~\ref{tbl:main_results} shows the main results of our experiments with task augmentation and self-training. Below, we first provide an overview of these results before analyzing them in more detail.
\vspace{-1mm}
\paragraph{Baselines:} \ssc{LMFT} is not always helpful and can even hurt performance (e.g., on \ssc{QNLI}, a task built from Wikipedia, which is part of \ssc{BERT}'s pre-training data). \citet{JDu20} also observe a decrease in performance when using \ssc{LMFT} with task-specific in-domain unlabeled data retrieved from Web data. \ssc{ITFT$_{\smallsup{MNLI}}$} significantly outperforms \ssc{LMFT} in many cases, particularly on target tasks closely related to \ssc{MNLI}.
\vspace{-1mm}
\paragraph{Task augmentation significantly improves results on downstream tasks: }
The first three blocks of Table~\ref{tbl:main_results} show the results for \ssc{TA}, which improves almost all target tasks across all three data regimes. \ssc{TA} even improves results on \ssc{SNLI} in the \ssc{Full} regime, where there is a large amount of labeled data available (570K examples). Changing the data regimes significantly impacts the average absolute performance gain over the vanilla \bertlarge\ \text{ across} target tasks, which is lowest in the \ssc{Full} regime (+2.7\%) and highest in 
the \ssc{Few-Shot} regime (+13.0\%). \ssc{SNLI} (+41.7\%) and \ssc{RTE} (+23.9\%) benefit the most from \ssc{TA} in the \ssc{Few-Shot} regime. \ssc{TA} also significantly outperforms both \ssc{LMFT} and \ssc{ITFT$_{\smallsup{MNLI}}$}, particularly in the low-data regimes (+16.4\% and +4.8\%, respectively).

\paragraph{Adding self-training further boosts downstream performance when task-specific unlabeled examples are available:} The third block of Table~\ref{tbl:main_results} shows that in the \ssc{Few-Shot} regime, adding \ssc{ST} to \ssc{TA}, which results in \strata, further boosts downstream performance. In particular, \strata performs the best across target tasks, achieving up to +44.2\% absolute improvement on \ssc{SNLI} over \bertlarge\.. Overall, \strata provides an average absolute performance gain of +20.9\% and +18.4\% for \bertbase\  \text{ and} \bertlarge\ , respectively. Using \ssc{ST} alone also leads to large improvements over the vanilla \ssc{BERT} models; however, the performance gain largely depends on the target task.

\paragraph{Using a stronger base model leads to better self-training results:}
Our experiment results show that self-training is complementary to different \ssc{BERT} models across target tasks---the stronger the \ssc{BERT} base model, the better self-training results. 
\ssc{BERT + TA} yields better self-training results than \ssc{BERT + ITFT$_{\smallsup{MNLI}}$}, and both are better than the vanilla \ssc{BERT}.
Combinations of \bertlarge\ \text{ and} \ssc{ST} typically outperform that of \bertbase\ \text{ and} \ssc{ST}. Interestingly, \bertlarge\ + \ssc{ST} is competitive with \bertlarge\ + \strata on several tasks (e.g., \ssc{QQP} and \ssc{QNLI}), and this does not hold for \bertbase\..
\vspace{-1mm}
\paragraph{Comparison to recent published work:}
\begin{table}[t]
\centering
\begin{adjustbox}{max width=0.40\textwidth}
\begin{tabular}{ l l l l}
\toprule
\textbf{Model} & \sc{SST-2} & \sc{SST-5} & \sc{CR} \\
\midrule
\midrule
\multicolumn{4}{l}{\emph{Ours (8 examples per class)}}\\
\lbertbase\ & \textcolor{black}{69.8$_{\smallsup{6.5}}$} & \textcolor{black}{32.8$_{\smallsup{2.0}}$} & \textcolor{black}{73.1$_{\smallsup{0.5}}$} \\
\hspace{10pt}\text{+ \sc{TA}} & \textcolor{black}{85.5$_{\smallsup{0.6}}$} & \textcolor{black}{41.0$_{\smallsup{0.8}}$} & \textcolor{black}{88.7$_{\smallsup{0.2}}$} \\
\hspace{10pt}\text{+ \sc{ST}} &
\textcolor{black}{74.9$_{\smallsup{9.0}}$} &
\textcolor{black}{38.3$_{\smallsup{0.8}}$} & \textcolor{black}{85.6$_{\smallsup{1.8}}$} \\
\hspace{10pt}\text{+ \lstrata} & \textcolor{black}{\textbf{90.8}$_{\smallsup{0.6}}$} & \textcolor{black}{\textbf{43.1}$_{\smallsup{1.1}}$} & \textcolor{black}{\textbf{91.4}$_{\smallsup{0.2}}$} \\
\midrule
\lbertlarge\ & \textcolor{black}{75.6$_{\smallsup{3.3}}$} & \textcolor{black}{36.6$_{\smallsup{0.4}}$} & \textcolor{black}{79.3$_{\smallsup{0.7}}$} \\
\hspace{10pt}\text{+ \sc{TA}} & \textcolor{black}{87.3$_{\smallsup{0.3}}$} & \textcolor{black}{41.7$_{\smallsup{1.1}}$} & \textcolor{black}{90.0$_{\smallsup{0.4}}$} \\
\hspace{10pt}\text{+ \sc{ST}} &
\textcolor{black}{90.6$_{\smallsup{0.3}}$} &
\textcolor{black}{43.8$_{\smallsup{0.4}}$} & \textcolor{black}{89.0$_{\smallsup{1.1}}$} \\
\hspace{10pt}\text{+ \lstrata} &
\textcolor{black}{\textbf{92.4}$_{\smallsup{0.1}}$} &
\textcolor{black}{\textbf{45.5}$_{\smallsup{0.7}}$} & \textcolor{black}{\textbf{90.6}$_{\smallsup{0.0}}$} \\
\midrule
\midrule
\multicolumn{4}{l}{\emph{\citet{JDu20} (20 examples per class)}} \\
\lrobertalarge\ & \textcolor{black}{83.6$_{\smallsup{2.7}}$} &
\textcolor{black}{42.3$_{\smallsup{1.6}}$} &
\textcolor{black}{88.9$_{\smallsup{1.7}}$} \\
\hspace{10pt}\text{+ \sc{SentAug-ST}}  & \textcolor{black}{\textbf{86.7}$_{\smallsup{2.3}}$} &
\textcolor{black}{\textbf{44.4}$_{\smallsup{1.0}}$} &
\textcolor{black}{\textbf{89.7}$_{\smallsup{2.0}}$} \\
\bottomrule
\end{tabular}
\end{adjustbox}
\caption{Compared to~\citet{JDu20}, our approach leads to better downstream performance, despite using a weaker base model (\ssc{BERT} vs. \ssc{RoBERTa}) and with less labeled examples.}
\label{tbl:8_t3}
\vspace{-2mm}
\end{table}

The last three rows of Table~\ref{tbl:main_results} and the last two rows of Table~\ref{tbl:8_t3} show results from recent published work.\footnote{While \citet{SWang21} report results for \fssc{LM-BFF} and \fssc{EFL} across 5 random data subsets using a fixed set of seeds, \citet{JDu20} tried 10 seeds for each of their 5 random data subsets and report the mean of the top 3 seeds. To be more comparable to \cite{JDu20}, we report the mean of our top 3 random seeds in Table~\ref{tbl:8_t3}.} Broadly, our methods lead to better performance compared to these approaches. However, due to differences in evaluation methodology (e.g., models, training/development data subsets, number of random restarts, and other factors), we refrain from explicitly ranking the %
approaches.

\section{Analysis of few-shot learning results}
\begin{figure}[t]
\centering
\includegraphics[width=0.48\textwidth]{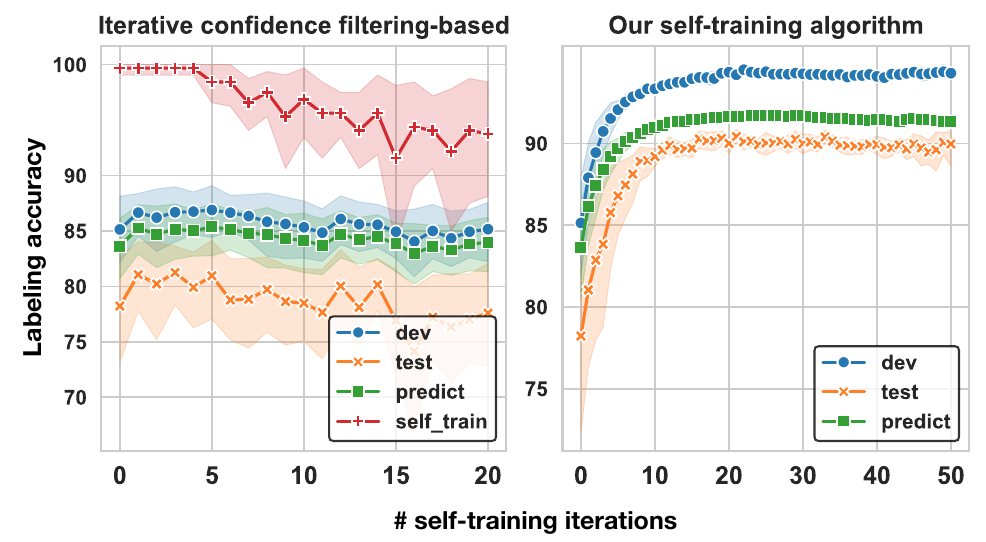}
\caption{On the \ssc{SST-2} sentiment dataset, traditional confidence filtering-based self-training (left) yields poor results compared to our approach, which trains on all pseudo-labels at each iteration (right).}
\label{fig:pseudo_labling}
\vspace*{-2mm}
\end{figure}

Having established the effectiveness of both task augmentation and self-training in the few-shot setting, we conduct a series of analysis experiments in this section to explore the source of the observed improvements.
\paragraph{Sample efficiency with \bstrata:}
Figure~\ref{fig:sample_efficiency} illustrates how our \strata approach improves sample efficiency as the number of examples per class increases. For the \ssc{SST-2} sentiment dataset, despite using only $K=8$ training examples per class%
, \strata has already nearly saturated its performance, achieving results competitive with standard fine-tuning over the whole dataset of 67K labeled examples. On the harder task of \ssc{SciTail}, \strata continues to improve as $K$ increases, and surpasses the performance of standard fine-tuning with the whole dataset of 27K labeled examples at $K = 512$.
\paragraph{\bstrata improves a randomly-initialized base model:}
\begin{table}[t]
\centering
\small
\begin{adjustbox}{max width=0.40\textwidth}
\begin{tabular}{ l l l}
\toprule
\textbf{Model} & \sc{SST-2} & \sc{SciTail} \\
\midrule
\midrule
\lrandbase\ &
\textcolor{black}{50.0$_{\smallsup{1.6}}$} &
\textcolor{black}{50.7$_{\smallsup{2.4}}$} \\
\hspace{10pt}\text{+ \lstrata} &
\textcolor{black}{\textbf{78.6}$_{\smallsup{0.9}}$} & \textcolor{black}{\textbf{64.4}$_{\smallsup{3.1}}$} \\
\midrule
\lbertbase\ &
\textcolor{black}{59.1$_{\smallsup{8.4}}$} &
\textcolor{black}{67.1$_{\smallsup{6.6}}$} \\
\hspace{10pt}\text{+ \lstrata} &
\textcolor{black}{\textbf{90.1}$_{\smallsup{0.8}}$} &
\textcolor{black}{\textbf{86.3}$_{\smallsup{3.5}}$}\\
\midrule
\lbertlarge\ &
\textcolor{black}{66.1$_{\smallsup{8.7}}$} &
\textcolor{black}{68.8$_{\smallsup{9.5}}$} \\
\hspace{10pt}\text{+ \lstrata} &
\textcolor{black}{\textbf{91.7}$_{\smallsup{0.7}}$} &
\textcolor{black}{\textbf{87.3}$_{\smallsup{2.9}}$}\\
\bottomrule
\end{tabular}
\end{adjustbox}
\caption{Our approach yields improvements even when starting with a randomly-initialized model, but pre-training helps considerably.}
\label{tbl:modeling_capability}
\vspace{-2mm}
\end{table}
Table~\ref{tbl:modeling_capability} shows that our \strata approach does not require a powerful pre-trained base model to exhibit improvements:  when applied to a randomly initialized Transfomer model (\randbase\ ) with the same architecture as \bertbase\ ,  \randbase\ + \strata outperforms the vanilla \bertbase\ \text{ by} a large margin on \ssc{SST-2}, while being competitive on \ssc{SciTail}. Additionally, \bertbase\ + \strata substantially outperforms the vanilla \bertlarge \text{ by} 24\% and 17.5\% on \ssc{SST-2} and \ssc{SciTail}, respectively. 
\paragraph{Self-training on a broad distribution of pseudo-labeled data:}
Previous self-training algorithms~\cite{CRosenberg05,DMcClosky06,KSohn20,JDu20} typically add a small set of unlabeled examples with the highest-confidence pseudo labels to the labeled data set $\bmmc{L}$ at each iteration. In contrast, our approach adds \emph{all} pseudo-labeled examples to $\bmmc{L}$ at every iteration regardless of confidence. We compare the two approaches in Figure~\ref{fig:pseudo_labling}, which shows the labeling accuracy (\% of unlabeled examples that are labeled correctly) on the development set (\textcolor{myblue}{\textbf{dev}}), the test set (\textcolor{myorange}{\textbf{test}}), and the unlabeled data pool (\textcolor{mygreen}{\textbf{predict}}) of the SST-2 sentiment dataset. In the iterative confidence filtering-based approach (left plot), a fixed number (in this plot, 32) of most confidently labeled examples are added to the labeled set $\bmmc{L}$ at each iteration (the \textcolor{myred}{\textbf{self-train}} line shows the labeling accuracy of these examples); once they have been added, they are not removed, and this process is repeated until the unlabeled \text{set } $\bmmc{U}$ is exhausted. As can be seen, this approach works well for the several first self-training iterations (3-5), but then labeling accuracy begins to degrade. %
In contrast, our algorithm (right plot) gradually and consistently improves labeling accuracy before converging at some iteration. %
These results suggest that strong base models benefit from including even significantly noisy pseudo-labels in self-training, as opposed to training on a narrow distribution of high-confidence predictions.
\paragraph{Does self-training work with out-of-domain/distribution (\bssc{OOD}) unlabeled examples?}
\label{sec:ood_unlabeled_data}
We investigate this question by applying self-training on top of \bertbase\ + \ssc{TA}. We consider \textsc{source $\rightarrow$ target} task pairs where training data from the source task without ground-truth labels is used as \ssc{OOD} unlabeled data for the target task. We experiment with several task pairs, including \ssc{MNLI} $\rightarrow$ \ssc{SciTail}, \ssc{SST-2} $\rightarrow$ \ssc{CR}, \ssc{QQP} $\rightarrow$ \ssc{MRPC}, and \ssc{MNLI} $\rightarrow$ \ssc{RTE}. As shown in Table~\ref{tbl:ood_st}, self-training with \ssc{OOD} unlabeled examples (\ssc{ST$_{\smallsup{OUT}}$}) is also helpful, offering an average absolute performance gain of +3.5\% over the strong \bertbase\ + \ssc{TA} baseline. However, using \ssc{OOD} unlabeled examples typically leads to worse self-training results compared to using in-domain unlabeled examples (\ssc{ST$_{\smallsup{IN}}$}), except for the case \ssc{MNLI} $\rightarrow$ \ssc{RTE}, and combining the two types of unlabeled examples (\ssc{ST$_{\smallsup{IN + OUT}}$}) does not bring further improvements over \ssc{ST$_{\smallsup{IN}}$}. 
\begin{table}[t]
\centering
\small
\begin{adjustbox}{max width=0.48\textwidth}
\begin{tabular}{ l l l l l l}
\toprule
\textbf{Model} & \sc{SciTail} & \sc{CR} & \sc{MRPC} & \sc{RTE} \\
\midrule
\midrule
\bertbase\ & \textcolor{black}{67.1$_{\smallsup{6.6}}$} & \textcolor{black}{65.2$_{\smallsup{8.2}}$} & \textcolor{black}{72.4$_{\smallsup{10.2}}$} & \textcolor{black}{51.4$_{\smallsup{2.5}}$} \\
\midrule
\bertbase\ \ssc{+ TA} & \textcolor{black}{78.5$_{\smallsup{3.2}}$} & \textcolor{black}{86.5$_{\smallsup{2.2}}$} & \textcolor{black}{74.5$_{\smallsup{6.5}}$} & \textcolor{black}{67.6$_{\smallsup{7.1}}$} \\
\hspace{10pt}\text{+ \ssc{ST$_{\smallsup{IN}}$}} & \textcolor{black}{\textbf{86.3}$_{\smallsup{3.5}}$} & \textcolor{black}{\textbf{90.5}$_{\smallsup{0.8}}$} & \textcolor{black}{\textbf{81.0}$_{\smallsup{0.8}}$} & \textcolor{black}{70.6$_{\smallsup{2.4}}$} \\
\hspace{10pt}\text{+ \ssc{ST$_{\smallsup{OUT}}$}} &
\textcolor{black}{81.4$_{\smallsup{3.7}}$} &
\textcolor{black}{88.3$_{\smallsup{1.9}}$} &
\textcolor{black}{80.3$_{\smallsup{1.9}}$} & \textcolor{black}{\textbf{71.2}$_{\smallsup{3.2}}$} \\
\hspace{10pt}\text{+ \ssc{ST$_{\smallsup{IN + OUT}}$}} &
\textcolor{black}{82.6$_{\smallsup{2.6}}$} &
\textcolor{black}{88.3$_{\smallsup{1.5}}$} &
\textcolor{black}{80.2$_{\smallsup{1.1}}$} &  \textcolor{black}{69.9$_{\smallsup{4.0}}$} \\
\bottomrule
\end{tabular}
\end{adjustbox}
\caption{Self-training with out-of-domain unlabeled examples also results in improvements, but using in-domain data works significantly better. }
\label{tbl:ood_st}
\end{table}

\begin{table}[t]
\centering
\small
\begin{adjustbox}{max width=0.40\textwidth}
\begin{tabular}{ l l l}
\toprule
\textbf{Model} & \sc{SST-2} & \sc{SciTail} \\
\midrule
\midrule
\bertbase\ &
\textcolor{black}{58.8$_{\smallsup{8.4}}$} ($\downarrow$ 0.3) &
\textcolor{black}{61.5$_{\smallsup{5.4}}$} ($\downarrow$ 5.6) \\
\hspace{10pt}\text{+ \ssc{LMFT}} &
\textcolor{black}{64.0$_{\smallsup{8.1}}$} ($\downarrow$ 0.9) & \textcolor{black}{59.3$_{\smallsup{5.6}}$} ($\downarrow$ 4.7) \\
\hspace{10pt}\text{+ \ssc{ITFT$_{\smallsup{MNLI}}$}} &
\textcolor{black}{76.5$_{\smallsup{7.2}}$} ($\downarrow$ 0.3) & \textcolor{black}{76.2$_{\smallsup{5.4}}$} ($\uparrow$ 0.4)\\
\hspace{10pt}\text{+ \ssc{TA}} &
\textcolor{black}{79.8$_{\smallsup{6.3}}$} ($\downarrow$ 0.5) & \textcolor{black}{77.8$_{\smallsup{3.3}}$} ($\downarrow$ 0.7) \\
\hspace{10pt}\text{+ \strata} &
\textcolor{black}{\textbf{86.6}$_{\smallsup{2.6}}$} ($\downarrow$ 3.5)
& \textcolor{black}{\textbf{80.6}$_{\smallsup{3.0}}$} ($\downarrow$ 5.7) \\
\bottomrule
\end{tabular}
\end{adjustbox}
\caption{In a realistic evaluation without a development set, our \strata approach still leads to significant improvements on top of \bertbase\ . In parentheses, we show the absolute increase ($\uparrow$) or decrease ($\downarrow$) in performance compared to the same method used with a development set.} 
\label{tbl:real_eval}
\vspace{-2mm}
\end{table}

\paragraph{Towards realistic evaluation in few-shot learning:} In real-world low-resource scenarios, it is often impractical to rely on a development set~\cite{AOliver18,KKann19}. With so little  data, it may be more effective to use all labeled data for training. To examine the applicability of our methods to this real-world setting, here we consider an evaluation that does not make use of a development set. Rather than using early stopping, we fine-tune each model for a fixed number of 512 steps. We checkpoint every 30 steps and evaluate a single model obtained by averaging the last 5 model checkpoints. For self-training, we perform a fixed number of 30 self-training iterations, each following the same fine-tuning procedure.

Table~\ref{tbl:real_eval} summarizes our results. Broadly, all models perform worse in this setting than when a development set is available. Our \strata approach still provides significant improvements over \bertbase\ , but much worse than the same method used with a development set. We conjecture that this is because without a development set, the model achieves somewhat lower accuracy in each self-training iteration, and these errors compound through later iterations.

\section{Related Work}
\label{section:related_work}
\paragraph{Improving language model fine-tuning:}
Fine-tuning has been the most common approach for applying pre-trained language models to downstream tasks. However, it typically requires a target dataset of thousands to tens of thousands of examples to work well~\cite{DYogatama19,TBrown20}. 
Many methods have been proposed to improve performance and stability of pre-trained language models on small datasets, including language model fine-tuning on unlabeled data from the target domain~\cite{JHoward18,SGururangan20}, intermediate-task fine-tuning~\cite{JPhang19}, multi-task pre-finetuning~\cite{AAghajanyan21a}, better design choices and training strategies~\cite{MMosbach21,TZhang21}, and regularization-oriented techniques~\cite{HJiang20,AAghajanyan21b}. More related to our work is research on intermediate-task training that makes use of data-rich tasks~\cite{JPhang19}, tasks that require complex reasoning and inference~\cite{YPruksachatkun20}, and beneficial relationships among tasks~\cite{TVu20,TVu21}. 
\vspace{-1mm}
\paragraph{Few-shot learning:} Our work also relates to research in few-shot learning. %
In previous work, fine-tuning is combined with other learning strategies to improve few-shot performance, including consistency training~\cite{QXie20a}, meta-learning~\cite{TBansal20}, self-training~\cite{JDu20,ZSun20}, and contrastive learning~\cite{BGunel21}. Other work has focused on prompt-based/entailment-based few-shot learning approaches~\cite{TBrown20,TSchick20,TGao20,DTam21,SWang21}. Notably, ~\citet{TBrown20} demonstrate remarkable few-shot learning performance with a single \textit{frozen} \ssc{GPT-3} model, although its performance still lags far behind state-of-the-art fine-tuning results.
\vspace{-1mm}
\paragraph{Generative data augmentation:}
Recent work explores the generation capabilities of large-scale generative language models, such as \ssc{GPT-2}~\cite{ARadford19} and \ssc{T5}~\cite{CRaffel20}, to generate synthetic training data for different tasks, including text classification~\cite{AAnaby-Tavor20,KLee21,TSchick21}, question answering~\cite{RPuri20}, and commonsense reasoning~\cite{YYang20}. \citet{YYang20} show that such a generative approach consistently outperforms previous data augmentation methods based on back-translation~\cite{RSennrich16,QXie20a}.
\vspace{-1mm}
\paragraph{Semi-supervised learning: } Another area upon which our work builds is semi-supervised learning (\ssc{SSL}). Recent work has combined self-training with other techniques, e.g., noise injection~\cite{JHe20,QXie20b}, consistency regularization and pseudo-labeling~\cite{KSohn20}, to develop powerful \ssc{SSL} algorithms. ~\citet{JDu20} show that self-training improves upon language model pre-training. %
\paragraph{Data augmentation for self-training: } There has been interest in data augmentation methods for self-training, where task-specific in-domain data is either retrieved from a large bank of Web data~\cite{JDu20} or synthesized by training task-specific generative models~\cite{XHe21}. Unlike these approaches, we use a single \ssc{NLI} data generator to produce in-domain \ssc{NLI} training examples for all tasks. Additionally, we demonstrate the importance of self-training on a broad distribution of pseudo-labeled data. These approaches are complementary, and combining them is a promising direction for future work.

\section{Limitations \& Conclusion}
Task augmentation and self-training provide complementary ways to leverage task-specific unlabeled data for improved downstream performance. While task augmentation utilizes unlabeled texts to 
synthesize %
a large amount of in-domain data for an auxiliary training task, self-training uses a model's predictions on unlabeled examples to improve the model itself. When combining these methods 
in \strata, 
we are able to substantially improve sample efficiency across 12 \ssc{NLP} benchmark datasets. That said, each method has its own limitations. While our implementation uses \ssc{NLI} as an auxiliary task in task augmentation, there are target tasks for which \ssc{NLI} may not be helpful (e.g., on grammatical acceptability judgments, as shown in~\citet{AWang19c}). Additionally, other auxiliary tasks may increase improvements (e.g., \ssc{QNLI} benefits more from \ssc{QA} tasks~\cite{TVu20}). We leave exploration of other auxiliary tasks to future work. Finally, our self-training algorithm (like prior approaches) assumes access to task-specific unlabeled examples, which might be non-trivial to acquire for some applications.

\section*{Acknowledgments}
We thank David Berthelot, Colin Raffel, Kalpesh Krishna, Zi Yang, Jenny Lee, Guolong Su, and Nan Hua for useful discussions and valuable feedback at different stages of this project. We would also like to thank the anonymous reviewers, Kenton Lee, Zihang Dai, Ed H. Chi, Nader Akoury, Brendan O'Connor, Zhiyang Xu, Andrew Drozdov, and the rest of the UMass NLP group for their thoughtful comments and suggestions.
\clearpage
\newpage
\bibliographystyle{acl_natbib}
\bibliography{emnlp2021}
\clearpage
\newpage
\appendix

\section*{Appendices}

\section{Additional details for the datasets used in our study}
\label{appendix:a}
\begin{table*}[t!]
\centering
\small
\begin{adjustbox}{max width=\textwidth}
\begin{tabular}{llll}
\toprule
\textbf{Task} & \textbf{$\mid$ Train $\mid$} & \textbf{Task type} & \textbf{Domain} \\ [0.5ex] 
\multicolumn{4}{l}{\emph{text classification/regression}} \\ [0.5ex] 
\textsc{SNLI}~\cite{SBowman15} & 570K & \textsc{NLI} & misc.\\
\textsc{MNLI}~\cite{AWilliams18} & 393K & \textsc{NLI} & misc. %
\\
\textsc{QQP}~\cite{SIyer17} & 364K & paraphrase identification & social \textsc{QA} \\
\textsc{QNLI}~\cite{AWang19} & 105K & \textsc{QA-NLI} & Wikipedia \\
\textsc{SST-2}~\cite{RSocher13} &  67K & sentiment analysis & movie reviews\\
\textsc{SciTail}~\cite{TKhot18} & 27K & \textsc{NLI} & science \textsc{QA}\\
\textsc{SST-5}~\cite{RSocher13} & 8.5K & sentiment analysis & movie reviews\\
\textsc{STS-B}~\cite{DCer17} & 7K & semantic similarity & misc. \\
\textsc{SICK-E}~\cite{MMarelli14} & 4.5K & \textsc{NLI} & misc. \\
\textsc{SICK-R}~\cite{MMarelli14} & 4.5K & semantic similarity & misc. \\
\textsc{CR}~\cite{MHu04} & 4K & sentiment analysis & product reviews \\
\textsc{MRPC}~\cite{WDolan05} & 3.7K & paraphrase identification & news \\
\textsc{RTE}~\cite[][et seq.]{IDagan06} & 2.5K & \textsc{NLI} & news, Wikipedia \\
\bottomrule
\end{tabular}
\end{adjustbox}
\caption{Datasets used in our experiments and their characteristics, sorted by training data set size.}
\label{tbl:dataset_characteristics}
\end{table*}
The datasets used in our experiments come from two common language understanding benchmarks: \ssc{GLUE}~\cite{AWang19} and \ssc{SentEval}~\cite{AConneau18}. See Table~\ref{tbl:dataset_characteristics} for details about dataset characteristics. We report F1 scores for \ssc{QQP} and \ssc{MRPC}, Spearman correlations for \ssc{STS-B} and \ssc{SICK-R}, and accuracy scores for the other tasks.

\section{Example outputs of our \bssc{NLI} data generator}
\label{appendix:b}
Table~\ref{tbl:model_outputs} shows example outputs of our \ssc{NLI} data generator using unlabeled texts from \ssc{SST-2} and \ssc{SciTail}.
\newpage
\begin{table*}[t!]
\centering
\small
\begin{adjustbox}{max width=\textwidth}
\begin{tabular}{l l}
\toprule
\multicolumn{2}{l}{\emph{unlabeled texts from \ssc{SST-2}}}\\[0.5ex] 
\midrule
\multicolumn{2}{l}{\textbf{Input:} a cast of a-list brit actors} \\	
\hspace{10pt}\text{+ \textbf{entailment}}	& $\rightarrow$ \textbf{Output:} the cast includes british actors. \\
\hspace{10pt}\text{+ \textbf{neutral}}	& $\rightarrow$ \textbf{Output:} the cast includes a-list brit actors such as nicole kidman and charlie chaplin. \\
\hspace{10pt}\text{+ \textbf{contradiction}} &	$\rightarrow$ \textbf{Output:} the cast was all new and unknown. \\
\midrule
\multicolumn{2}{l}{\textbf{Input:} both heartbreaking and heartwarming ... just a simple fable done in an artless sytle , but it 's tremendously moving} \\	
\hspace{10pt}\text{+ \textbf{entailment}} & $\rightarrow$ \textbf{Output:} the story is moving and touching. \\	
\hspace{10pt}\text{+ \textbf{neutral}}	& $\rightarrow$ \textbf{Output:} the artless style used to make the fable was a favorite of many. \\	
\hspace{10pt}\text{+ \textbf{contradiction}} & $\rightarrow$ \textbf{Output:} the fable is artful and sophisticated but it is also heartbreaking. \\
\midrule
\multicolumn{2}{l}{\textbf{Input:} gaping plot holes sink this ` sub ' - standard thriller and drag audience enthusiasm to crush depth} \\
\hspace{10pt}\text{+ \textbf{entailment}}	& $\rightarrow$ \textbf{Output:} this movie's plot holes seriously affect the enjoyment of the movie. \\
\hspace{10pt}\text{+ \textbf{neutral}}	& $\rightarrow$ \textbf{Output:} the movies pace is too slow for most audiences. \\
\hspace{10pt}\text{+ \textbf{contradiction}} & $\rightarrow$ \textbf{Output:} the movie is a fun, easy romp of a thriller. \\
\midrule
\multicolumn{2}{l}{\textbf{Input:} putting the primitive murderer inside a high-tech space station unleashes a pandora 's box of special effects that run the gamut } \\
\multicolumn{2}{l}{from cheesy to cheesier to cheesiest} \\
\hspace{10pt}\text{+ \textbf{entailment}} & $\rightarrow$ \textbf{Output:} the special effects are all over the place. \\
\hspace{10pt}\text{+ \textbf{entailment}} & $\rightarrow$ \textbf{Output:} putting a murderer in a space station leads to all sorts of cheesy special effects. \\
\hspace{10pt}\text{+ \textbf{entailment}} & $\rightarrow$ \textbf{Output:} the murderer was inside a space station. \\
\midrule
\multicolumn{2}{l}{\textbf{Input:} vera has created a provocative , absorbing drama that reveals the curse of a self-hatred instilled by rigid social mores . } \\	
\hspace{10pt}\text{+ \textbf{neutral}} & $\rightarrow$ \textbf{Output:} vera has created a provocative drama that is absorbing and has a great cast. \\
\hspace{10pt}\text{+ \textbf{neutral}} & $\rightarrow$ \textbf{Output:} vera has created a provocative drama that shows the curse of self-hatred and shows the bonds \\
& between women. \\
\hspace{10pt}\text{+ \textbf{neutral}} & $\rightarrow$ \textbf{Output:} vera's drama, which has been loved by millions, is about the curse of self-hatred. \\
\midrule
\multicolumn{2}{l}{\textbf{Input:}  filmmaking from one of french cinema 's master craftsmen } \\	
\hspace{10pt}\text{+ \textbf{contradiction}} & $\rightarrow$ \textbf{Output:} the filmmaking is from a newcomer to the french cinema. \\
\hspace{10pt}\text{+ \textbf{contradiction}} & $\rightarrow$ \textbf{Output:} the filmmaking is from a new blood in the american cinema. \\
\hspace{10pt}\text{+ \textbf{contradiction}} & $\rightarrow$ \textbf{Output:} the filmmaking of one of italian cinema's master craftsmen \\
\midrule
\midrule
\multicolumn{2}{l}{\emph{unlabeled texts from \ssc{SciTail}}}\\[0.5ex] 
\midrule
\multicolumn{2}{l}{\textbf{Input:} Surface tension is responsible for the fact that small insects can walk on water.} \\	
\hspace{10pt}\text{+ \textbf{entailment}} & $\rightarrow$ \textbf{Output:} small insects can walk on water due to surface tension. \\
\hspace{10pt}\text{+ \textbf{neutral}} & $\rightarrow$ \textbf{Output:} it's difficult for large insects to walk on water. \\
\hspace{10pt}\text{+ \textbf{contradiction}} & $\rightarrow$ \textbf{Output:} small insects cannot walk on water because of surface tension.\\
\midrule
\multicolumn{2}{l}{\textbf{Input:} Tamiflu inhibits spread of virus.	} \\
\hspace{10pt}\text{+ \textbf{entailment}} & $\rightarrow$ \textbf{Output:} the virus is less spread when tamiflu is used. \\
\hspace{10pt}\text{+ \textbf{neutral}} & $\rightarrow$ \textbf{Output:} tamiflu is a good way to treat hiv. \\
\hspace{10pt}\text{+ \textbf{contradiction}} & $\rightarrow$ \textbf{Output:} tamiflu promotes viral spread. \\
\midrule
\multicolumn{2}{l}{\textbf{Input:} Animals can transport the waste material known as urea in the circulatory system and store it safely at high concentrations.	} \\
\hspace{10pt}\text{+ \textbf{entailment}} & $\rightarrow$ \textbf{Output:} animals can transport waste material in their circulatory systems. \\
\hspace{10pt}\text{+ \textbf{neutral}} & $\rightarrow$ \textbf{Output:} animals can store urea in their bloodstreams for up to a year. \\
\hspace{10pt}\text{+ \textbf{contradiction}} & $\rightarrow$ \textbf{Output:} only plants can transport the waste material known as urea in their circulatory systems. \\
\midrule
\multicolumn{2}{l}{\textbf{Input:} A number of applications of biomass, wind, hydropower and solar thermal are presently cost competitive with fossil fuels. } \\
\hspace{10pt}\text{+ \textbf{entailment}} & $\rightarrow$ \textbf{Output:} many biomass applications are now cost competitive with fossil fuels. \\
\hspace{10pt}\text{+ \textbf{entailment}} & $\rightarrow$ \textbf{Output:} many alternatives to fossil fuels are cost competitive. \\
\hspace{10pt}\text{+ \textbf{entailment}} & $\rightarrow$ \textbf{Output:} some biofuels are now cost competitive with fossil fuels. \\
\midrule
\multicolumn{2}{l}{\textbf{Input:} A cell wall is not present in animal cells. } \\
\hspace{10pt}\text{+ \textbf{neutral}} & $\rightarrow$ \textbf{Output:} in contrast, plant cells have a cell wall. \\
\hspace{10pt}\text{+ \textbf{neutral}} & $\rightarrow$ \textbf{Output:} in addition to not having a cell wall, animal cells also lack mitochondria. \\
\hspace{10pt}\text{+ \textbf{neutral}} & $\rightarrow$ \textbf{Output:} in animal cells, there is no cell wall, said the biologist. \\
\midrule
\multicolumn{2}{l}{\textbf{Input:} A bathymetric map can show the features of the bottom of a body of water.	} \\
\hspace{10pt}\text{+ \textbf{contradiction}} & $\rightarrow$ \textbf{Output:} a bathymetric map shows the top of a body of water. \\
\hspace{10pt}\text{+ \textbf{contradiction}} & $\rightarrow$ \textbf{Output:} a bathymetric map shows the features of the sky. \\
\hspace{10pt}\text{+ \textbf{contradiction}} & $\rightarrow$ \textbf{Output:} a bathymetric map shows what is on the surface of  the water. \\
\bottomrule
\end{tabular}
\end{adjustbox}
\caption{Example outputs of our \ssc{NLI} data generator using unlabeled texts from \ssc{SST-2} and \ssc{SciTail}.}
\label{tbl:model_outputs}
\end{table*}

\section{Full experiment results across data regimes}
\label{appendix:d}
We report the full results for our experiments in the \ssc{Full}, \ssc{Limited}, and \ssc{Few-shot} data regimes in Table~\ref{tbl:full}, Table~\ref{tbl:1024}, and Table~\ref{tbl:8}, respectively.
\begin{table*}[t]
\centering
\begin{adjustbox}{max width=\textwidth}
\begin{tabular}{ l l l l l  l l l l  l l l l}
\toprule
\textbf{Model} & \sc{SNLI} & \sc{QQP} & \sc{QNLI} & \sc{SST-2} & \sc{SciTail} & \sc{SST-5} & \sc{STS-B} & \sc{SICK-E} & \sc{SICK-R} & \sc{CR} & \sc{MRPC} & \sc{RTE} \\
\midrule
\midrule
\lbertbase\ & \textcolor{black}{90.3} & \textcolor{black}{87.8} & \textcolor{black}{90.6} & \textcolor{black}{91.7} & \textcolor{black}{93.2} & \textcolor{black}{52.7$_{\smallsup{0.8}}$} & \textcolor{black}{88.9$_{\smallsup{0.3}}$} & \textcolor{black}{86.7$_{\smallsup{0.5}}$} & \textcolor{black}{82.9$_{\smallsup{0.5}}$} & \textcolor{black}{91.0$_{\smallsup{0.9}}$} & \textcolor{black}{87.9$_{\smallsup{1.0}}$} & \textcolor{black}{63.5$_{\smallsup{2.3}}$} \\
\hspace{10pt}$\text{+ \sc{LMFT}}$ &
\textcolor{black}{90.8} & \textcolor{black}{87.8} & \textcolor{black}{90.2} & \textcolor{black}{91.3} & \textcolor{black}{92.9} & \textcolor{black}{52.8$_{\smallsup{0.9}}$} & \textcolor{black}{89.3$_{\smallsup{0.3}}$} & \textcolor{black}{86.8$_{\smallsup{0.8}}$} & \textcolor{black}{82.7$_{\smallsup{0.5}}$} & \textcolor{black}{90.5$_{\smallsup{1.0}}$} & \textcolor{black}{87.9$_{\smallsup{0.6}}$} & \textcolor{black}{63.9$_{\smallsup{3.7}}$} \\
\hspace{10pt}\text{+ \sc{ITFT$_{\smallsup{MNLI}}$}} &
\textcolor{black}{91.0} & \textcolor{black}{87.7} & \textcolor{black}{90.3} & \textcolor{black}{93.0} & \textcolor{black}{95.8} & \textcolor{black}{53.8$_{\smallsup{0.8}}$} & \textcolor{black}{\textbf{90.1}$_{\smallsup{0.1}}$} & \textcolor{black}{89.5$_{\smallsup{0.3}}$} & \textcolor{black}{85.3$_{\smallsup{0.6}}$} & \textcolor{black}{91.7$_{\smallsup{0.7}}$} & \textcolor{black}{89.8$_{\smallsup{1.1}}$} & \textcolor{black}{78.1$_{\smallsup{1.9}}$} \\
\hspace{10pt}\text{+ \sc{TA}} & \textcolor{black}{\textbf{91.2}} & \textcolor{black}{\textbf{88.1}} & \textcolor{black}{\textbf{90.9}} & \textcolor{black}{\textbf{93.9}} & \textcolor{black}{\textbf{96.3}} & \textcolor{black}{\textbf{54.3}$_{\smallsup{0.9}}$} & \textcolor{black}{\textbf{90.1}$_{\smallsup{0.1}}$} &
\textcolor{black}{\textbf{90.1}$_{\smallsup{0.3}}$} & \textcolor{black}{\textbf{85.6}$_{\smallsup{0.3}}$} & \textcolor{black}{\textbf{92.2}$_{\smallsup{0.5}}$} & \textcolor{black}{\textbf{90.1}$_{\smallsup{0.8}}$} & \textcolor{black}{\textbf{79.3}$_{\smallsup{0.9}}$}
\\
\midrule
\lbertlarge\ &  \textcolor{black}{91.1} & \textcolor{black}{88.4} & \textcolor{black}{91.9} & \textcolor{black}{92.4} & \textcolor{black}{95.3} & \textcolor{black}{53.7$_{\smallsup{0.9}}$} & \textcolor{black}{89.6$_{\smallsup{0.2}}$} & \textcolor{black}{87.9$_{\smallsup{0.6}}$} & \textcolor{black}{84.4$_{\smallsup{0.4}}$} & \textcolor{black}{91.7$_{\smallsup{0.6}}$} & \textcolor{black}{89.0$_{\smallsup{0.8}}$} & \textcolor{black}{68.6$_{\smallsup{7.2}}$} \\
\hspace{10pt}$\text{+ \sc{LMFT}}$ & \textcolor{black}{91.0} &
\textcolor{black}{88.1} & \textcolor{black}{90.4} & \textcolor{black}{93.5} & \textcolor{black}{95.3} & \textcolor{black}{54.0$_{\smallsup{0.4}}$} & \textcolor{black}{89.5$_{\smallsup{0.2}}$} & \textcolor{black}{87.7$_{\smallsup{0.5}}$} & \textcolor{black}{84.0$_{\smallsup{0.5}}$} & \textcolor{black}{91.6$_{\smallsup{0.8}}$} & \textcolor{black}{89.5$_{\smallsup{1.0}}$} & \textcolor{black}{66.5$_{\smallsup{7.3}}$} \\
\hspace{10pt}\text{+ \sc{ITFT$_{\smallsup{MNLI}}$}} &
\textcolor{black}{91.1} & \textcolor{black}{88.2} & \textcolor{black}{91.6} & \textcolor{black}{93.5} & \textcolor{black}{96.5} & \textcolor{black}{54.0$_{\smallsup{0.8}}$} & \textcolor{black}{90.3$_{\smallsup{0.3}}$} & \textcolor{black}{89.9$_{\smallsup{0.2}}$} & \textcolor{black}{86.3$_{\smallsup{0.3}}$} & \textcolor{black}{92.0$_{\smallsup{0.6}}$} & \textcolor{black}{89.7$_{\smallsup{0.9}}$} & \textcolor{black}{82.3$_{\smallsup{1.4}}$} \\
\hspace{10pt}\text{+ \sc{TA}} & \textcolor{black}{\textbf{91.9}} & \textcolor{black}{\textbf{88.5}} & \textcolor{black}{\textbf{92.5}} & \textcolor{black}{\textbf{94.7}} & \textcolor{black}{\textbf{96.9}} & \textcolor{black}{\textbf{55.7}$_{\smallsup{0.8}}$} & \textcolor{black}{\textbf{90.9}$_{\smallsup{0.2}}$} & \textcolor{black}{\textbf{90.7}$_{\smallsup{0.3}}$} & \textcolor{black}{\textbf{87.0}$_{\smallsup{0.3}}$} & \textcolor{black}{\textbf{93.3}$_{\smallsup{0.6}}$} \textcolor{black} & {\textbf{90.8}$_{\smallsup{0.7}}$} & \textcolor{black}{\textbf{83.8}$_{\smallsup{1.1}}$}\\
\bottomrule
\end{tabular}
\end{adjustbox}
\caption{Our experiment results in the \ssc{Full} data regime.}
\label{tbl:full}
\end{table*}

\begin{table*}[t]
\centering
\begin{adjustbox}{max width=\textwidth}
\begin{tabular}{ l l l l l  l l l l  l l l l}
\toprule
\textbf{Model} & \sc{SNLI} & \sc{QQP} & \sc{QNLI} & \sc{SST-2} & \sc{SciTail} & \sc{SST-5} & \sc{STS-B} & \sc{SICK-E} & \sc{SICK-R} & \sc{CR} & \sc{MRPC} & \sc{RTE} \\
\midrule
\midrule
\lbertbase\ & \textcolor{black}{71.7$_{\smallsup{1.1}}$} & \textcolor{black}{71.7$_{\smallsup{0.4}}$} & \textcolor{black}{78.7$_{\smallsup{0.7}}$} & \textcolor{black}{87.4$_{\smallsup{1.0}}$} & \textcolor{black}{88.3$_{\smallsup{1.2}}$} & \textcolor{black}{47.1$_{\smallsup{1.3}}$} & \textcolor{black}{86.8$_{\smallsup{0.6}}$} & \textcolor{black}{81.5$_{\smallsup{0.6}}$} & \textcolor{black}{76.7$_{\smallsup{0.8}}$} & \textcolor{black}{89.9$_{\smallsup{0.7}}$} & \textcolor{black}{83.9$_{\smallsup{1.1}}$} & \textcolor{black}{61.7$_{\smallsup{1.3}}$} \\
\hspace{10pt}$\text{+ \sc{LMFT}}$ & \textcolor{black}{73.4$_{\smallsup{2.1}}$} & \textcolor{black}{72.1$_{\smallsup{0.6}}$} & \textcolor{black}{76.3$_{\smallsup{3.1}}$} & \textcolor{black}{86.9$_{\smallsup{1.2}}$} & \textcolor{black}{88.4$_{\smallsup{1.4}}$} & \textcolor{black}{47.5$_{\smallsup{1.3}}$} & \textcolor{black}{87.2$_{\smallsup{0.7}}$} & \textcolor{black}{81.1$_{\smallsup{0.6}}$} & \textcolor{black}{75.9$_{\smallsup{0.7}}$} & \textcolor{black}{91.1$_{\smallsup{0.8}}$} & \textcolor{black}{84.4$_{\smallsup{0.6}}$} & \textcolor{black}{63.2$_{\smallsup{2.3}}$} \\
\hspace{10pt}\text{+ \sc{ITFT$_{\smallsup{MNLI}}$}} &
\textcolor{black}{82.9$_{\smallsup{0.3}}$} & \textcolor{black}{73.3$_{\smallsup{0.7}}$} & \textcolor{black}{81.6$_{\smallsup{0.9}}$} & \textcolor{black}{87.8$_{\smallsup{0.6}}$} & \textcolor{black}{90.3$_{\smallsup{1.1}}$} & \textcolor{black}{48.8$_{\smallsup{1.0}}$} & \textcolor{black}{\textbf{88.5}$_{\smallsup{0.3}}$} & \textcolor{black}{87.6$_{\smallsup{0.5}}$} & \textcolor{black}{81.7$_{\smallsup{0.8}}$} & \textcolor{black}{90.0$_{\smallsup{0.6}}$} & \textcolor{black}{87.0$_{\smallsup{0.9}}$} & \textcolor{black}{78.0$_{\smallsup{1.3}}$} \\
\hspace{10pt}\text{+ \sc{TA}} & \textcolor{black}{\textbf{85.7}$_{\smallsup{0.3}}$} & \textcolor{black}{\textbf{75.3}$_{\smallsup{0.4}}$} & \textcolor{black}{\textbf{82.5}$_{\smallsup{0.8}}$} & \textcolor{black}{\textbf{90.4}$_{\smallsup{0.7}}$} & \textcolor{black}{\textbf{90.7}$_{\smallsup{0.8}}$} & \textcolor{black}{\textbf{49.2}$_{\smallsup{1.3}}$} &
\textcolor{black}{\textbf{88.5}$_{\smallsup{0.3}}$} & \textcolor{black}{\textbf{88.5}$_{\smallsup{0.5}}$} & \textcolor{black}{\textbf{82.4}$_{\smallsup{0.6}}$} & \textcolor{black}{\textbf{91.4}$_{\smallsup{1.0}}$} & \textcolor{black}{\textbf{87.3}$_{\smallsup{0.7}}$} & \textcolor{black}{\textbf{78.7}$_{\smallsup{1.2}}$}
\\
\midrule
\lbertlarge\ & \textcolor{black}{77.4$_{\smallsup{0.6}}$} & \textcolor{black}{74.1$_{\smallsup{1.0}}$} & \textcolor{black}{81.7$_{\smallsup{0.9}}$} & \textcolor{black}{89.8$_{\smallsup{0.6}}$} & \textcolor{black}{90.9$_{\smallsup{0.7}}$} & \textcolor{black}{49.1$_{\smallsup{1.3}}$} & \textcolor{black}{88.2$_{\smallsup{0.4}}$} & \textcolor{black}{84.8$_{\smallsup{0.7}}$} & \textcolor{black}{80.2$_{\smallsup{0.4}}$} & \textcolor{black}{91.2$_{\smallsup{0.6}}$} & \textcolor{black}{85.7$_{\smallsup{1.7}}$} & \textcolor{black}{66.8$_{\smallsup{2.7}}$} \\
\hspace{10pt}$\text{+ \sc{LMFT}}$ &
\textcolor{black}{75.8$_{\smallsup{1.5}}$} & \textcolor{black}{71.6$_{\smallsup{0.5}}$} & \textcolor{black}{80.5$_{\smallsup{2.0}}$} & \textcolor{black}{88.9$_{\smallsup{0.8}}$} & \textcolor{black}{87.7$_{\smallsup{2.3}}$} & \textcolor{black}{49.2$_{\smallsup{3.1}}$} & \textcolor{black}{88.4$_{\smallsup{0.4}}$} & \textcolor{black}{83.2$_{\smallsup{0.6}}$} & \textcolor{black}{78.5$_{\smallsup{0.6}}$} & \textcolor{black}{90.9$_{\smallsup{0.7}}$} & \textcolor{black}{84.9$_{\smallsup{1.1}}$} & \textcolor{black}{65.2$_{\smallsup{3.4}}$} \\
\hspace{10pt}\text{+ \sc{ITFT$_{\smallsup{MNLI}}$}} &
\textcolor{black}{85.2$_{\smallsup{0.4}}$} & \textcolor{black}{74.0$_{\smallsup{0.5}}$} & \textcolor{black}{83.5$_{\smallsup{0.5}}$} & \textcolor{black}{90.0$_{\smallsup{0.8}}$} & \textcolor{black}{92.1$_{\smallsup{1.1}}$} & \textcolor{black}{49.4$_{\smallsup{1.2}}$} & \textcolor{black}{87.8$_{\smallsup{0.8}}$} & \textcolor{black}{88.8$_{\smallsup{0.5}}$} & \textcolor{black}{83.2$_{\smallsup{0.7}}$} & \textcolor{black}{91.3$_{\smallsup{0.7}}$} & \textcolor{black}{86.4$_{\smallsup{0.9}}$} & \textcolor{black}{81.1$_{\smallsup{1.3}}$} \\
\hspace{10pt}\text{+ \sc{TA}} & \textcolor{black}{\textbf{87.3}$_{\smallsup{0.3}}$} & \textcolor{black}{\textbf{75.7}$_{\smallsup{0.5}}$} & \textcolor{black}{\textbf{85.0}$_{\smallsup{0.5}}$} & \textcolor{black}{\textbf{91.7}$_{\smallsup{0.7}}$} & \textcolor{black}{\textbf{92.3}$_{\smallsup{1.1}}$} & \textcolor{black}{\textbf{51.4}$_{\smallsup{1.0}}$} & \textcolor{black}{\textbf{89.0}$_{\smallsup{0.6}}$} & \textcolor{black}{\textbf{89.4}$_{\smallsup{0.4}}$} & \textcolor{black}{\textbf{84.3}$_{\smallsup{0.4}}$} & \textcolor{black}{\textbf{92.6}$_{\smallsup{0.6}}$} & \textcolor{black}{\textbf{88.0}$_{\smallsup{0.8}}$} & \textcolor{black}{\textbf{82.9}$_{\smallsup{1.8}}$} \\
\bottomrule
\end{tabular}
\end{adjustbox}
\caption{Our experiment results in the \ssc{Limited} data regime.}
\label{tbl:1024}
\end{table*}

\begin{table*}[t]
\centering
\begin{adjustbox}{max width=0.99\textwidth}
\begin{tabular}{ l l l l l  l l l l  l l l l}
\toprule
\textbf{Model} & \sc{SNLI} & \sc{QQP} & \sc{QNLI} & \sc{SST-2} & \sc{SciTail} & \sc{SST-5} & \sc{STS-B} & \sc{SICK-E} & \sc{SICK-R} & \sc{CR} & \sc{MRPC} & \sc{RTE} \\
\midrule
\midrule
\lbertbase\ &
 \textcolor{black}{43.7$_{\smallsup{2.2}}$} & \textcolor{black}{55.9$_{\smallsup{6.5}}$} & \textcolor{black}{59.0$_{\smallsup{10.9}}$} & \textcolor{black}{59.1$_{\smallsup{8.4}}$} & \textcolor{black}{67.1$_{\smallsup{6.6}}$} & \textcolor{black}{30.5$_{\smallsup{2.0}}$} & \textcolor{black}{73.6$_{\smallsup{4.5}}$} & \textcolor{black}{61.3$_{\smallsup{4.1}}$} & \textcolor{black}{59.7$_{\smallsup{2.7}}$} & \textcolor{black}{65.2$_{\smallsup{8.2}}$} & \textcolor{black}{72.4$_{\smallsup{10.2}}$} & \textcolor{black}{51.4$_{\smallsup{2.5}}$} \\
\hspace{10pt}$\text{+ \sc{LMFT}}$ &
\textcolor{black}{45.2$_{\smallsup{3.9}}$} & \textcolor{black}{57.2$_{\smallsup{6.2}}$} & \textcolor{black}{57.6$_{\smallsup{9.1}}$} & \textcolor{black}{64.9$_{\smallsup{8.7}}$} & \textcolor{black}{64.0$_{\smallsup{8.0}}$} & \textcolor{black}{33.4$_{\smallsup{1.9}}$} & \textcolor{black}{75.4$_{\smallsup{4.4}}$} & \textcolor{black}{59.3$_{\smallsup{4.0}}$} & \textcolor{black}{58.3$_{\smallsup{2.0}}$} & \textcolor{black}{72.4$_{\smallsup{6.0}}$} & \textcolor{black}{73.9$_{\smallsup{8.6}}$} & \textcolor{black}{50.9$_{\smallsup{3.9}}$} \\
\hspace{10pt}\text{+ \sc{ITFT$_{\smallsup{MNLI}}$}} &
\textcolor{black}{75.2$_{\smallsup{5.7}}$} & \textcolor{black}{63.7$_{\smallsup{7.0}}$} & \textcolor{black}{62.8$_{\smallsup{5.1}}$} & \textcolor{black}{76.8$_{\smallsup{7.2}}$} & \textcolor{black}{75.8$_{\smallsup{5.6}}$} & \textcolor{black}{35.0$_{\smallsup{2.6}}$} & \textcolor{black}{80.2$_{\smallsup{1.1}}$} & \textcolor{black}{80.4$_{\smallsup{1.9}}$} & \textcolor{black}{73.5$_{\smallsup{2.7}}$} & \textcolor{black}{79.2$_{\smallsup{3.6}}$} & \textcolor{black}{74.3$_{\smallsup{8.0}}$} & \textcolor{black}{62.2$_{\smallsup{13.5}}$} \\
\hspace{10pt}\text{+ \sc{TA}} & \textcolor{black}{83.3$_{\smallsup{0.8}}$} & \textcolor{black}{68.7$_{\smallsup{1.5}}$} & \textcolor{black}{70.1$_{\smallsup{3.4}}$} & \textcolor{black}{80.3$_{\smallsup{6.6}}$} & \textcolor{black}{78.5$_{\smallsup{3.2}}$} & \textcolor{black}{37.4$_{\smallsup{3.0}}$} & \textcolor{black}{80.7$_{\smallsup{1.5}}$} & \textcolor{black}{81.1$_{\smallsup{2.4}}$} & \textcolor{black}{75.9$_{\smallsup{1.8}}$} & \textcolor{black}{86.5$_{\smallsup{2.2}}$} & \textcolor{black}{74.5$_{\smallsup{6.5}}$} & \textcolor{black}{67.6$_{\smallsup{7.1}}$} \\
\hspace{10pt}\text{+ \sc{ST}} &
\textcolor{black}{65.0$_{\smallsup{5.8}}$} & \textcolor{black}{69.9$_{\smallsup{5.9}}$} &
\textcolor{black}{71.6$_{\smallsup{11.3}}$} &
\textcolor{black}{62.7$_{\smallsup{10.4}}$} & \textcolor{black}{68.6$_{\smallsup{8.3}}$} & \textcolor{black}{33.9$_{\smallsup{3.5}}$} &
\textcolor{black}{80.5$_{\smallsup{2.2}}$} &
\textcolor{black}{68.1$_{\smallsup{4.5}}$} &
\textcolor{black}{64.0$_{\smallsup{2.4}}$} &
\textcolor{black}{78.2$_{\smallsup{6.3}}$} &
\textcolor{black}{80.5$_{\smallsup{1.8}}$} &
\textcolor{black}{50.7$_{\smallsup{3.1}}$} \\

\hspace{10pt}\text{+ \sc{ITFT$_{\smallsup{MNLI}}$} \text{+ ST}} &
\textcolor{black}{83.2$_{\smallsup{0.3}}$} &
\textcolor{black}{70.7$_{\smallsup{5.9}}$} &
\textcolor{black}{81.5$_{\smallsup{1.2}}$} &
\textcolor{black}{88.0$_{\smallsup{2.1}}$} & \textcolor{black}{83.7$_{\smallsup{4.4}}$} & \textcolor{black}{39.5$_{\smallsup{2.0}}$} &
\textcolor{black}{84.2$_{\smallsup{0.8}}$} &
\textcolor{black}{81.8$_{\smallsup{2.6}}$} & \textcolor{black}{75.8$_{\smallsup{2.2}}$} & \textcolor{black}{85.6$_{\smallsup{2.3}}$} &
\textcolor{black}{80.6$_{\smallsup{1.2}}$} &
\textcolor{black}{62.5$_{\smallsup{12.0}}$} \\
\hspace{10pt}\text{+ \lstrata} &
\textcolor{black}{\textbf{85.7}$_{\smallsup{0.2}}$} & \textcolor{black}{\textbf{74.5}$_{\smallsup{0.4}}$} &
\textcolor{black}{\textbf{82.1}$_{\smallsup{0.5}}$} &
\textcolor{black}{\textbf{90.1}$_{\smallsup{0.8}}$} & \textcolor{black}{\textbf{86.3}$_{\smallsup{3.5}}$} & \textcolor{black}{\textbf{41.3}$_{\smallsup{1.5}}$} &
\textcolor{black}{\textbf{84.7}$_{\smallsup{0.5}}$} &
\textcolor{black}{\textbf{84.9}$_{\smallsup{1.2}}$} & \textcolor{black}{\textbf{77.6}$_{\smallsup{1.6}}$} &
\textcolor{black}{\textbf{90.5}$_{\smallsup{0.8}}$} &
\textcolor{black}{\textbf{81.0}$_{\smallsup{0.8}}$} &
\textcolor{black}{\textbf{70.6}$_{\smallsup{2.4}}$} \\
\midrule
\lbertlarge\ &
\textcolor{black}{43.1$_{\smallsup{4.4}}$} & \textcolor{black}{58.5$_{\smallsup{4.7}}$} & \textcolor{black}{64.4$_{\smallsup{6.1}}$} & \textcolor{black}{66.1$_{\smallsup{8.7}}$} & \textcolor{black}{68.8$_{\smallsup{9.5}}$} & \textcolor{black}{35.2$_{\smallsup{1.3}}$} & \textcolor{black}{74.6$_{\smallsup{3.8}}$} & \textcolor{black}{66.5$_{\smallsup{4.5}}$} & \textcolor{black}{66.6$_{\smallsup{3.3}}$} & \textcolor{black}{72.0$_{\smallsup{6.0}}$} & \textcolor{black}{79.9$_{\smallsup{2.0}}$} & \textcolor{black}{53.1$_{\smallsup{3.3}}$} \\
\hspace{10pt}$\text{+ \sc{LMFT}}$ &
\textcolor{black}{39.6$_{\smallsup{2.6}}$} & \textcolor{black}{52.7$_{\smallsup{4.7}}$} & \textcolor{black}{52.2$_{\smallsup{1.6}}$} & \textcolor{black}{66.3$_{\smallsup{9.3}}$} & \textcolor{black}{66.4$_{\smallsup{10.6}}$} & \textcolor{black}{36.8$_{\smallsup{2.9}}$} & \textcolor{black}{75.4$_{\smallsup{9.4}}$} & \textcolor{black}{58.8$_{\smallsup{6.9}}$} & \textcolor{black}{51.6$_{\smallsup{7.0}}$} & \textcolor{black}{75.6$_{\smallsup{5.9}}$} & \textcolor{black}{80.5$_{\smallsup{2.4}}$} & \textcolor{black}{52.8$_{\smallsup{4.8}}$} \\
\hspace{10pt}\text{+ \sc{ITFT$_{\smallsup{MNLI}}$}}  & \textcolor{black}{79.9$_{\smallsup{3.1}}$} & \textcolor{black}{62.6$_{\smallsup{9.0}}$} & \textcolor{black}{64.5$_{\smallsup{4.4}}$} & \textcolor{black}{80.7$_{\smallsup{5.0}}$} & \textcolor{black}{72.3$_{\smallsup{11.2}}$} & \textcolor{black}{36.4$_{\smallsup{2.1}}$} & \textcolor{black}{75.5$_{\smallsup{4.0}}$} & \textcolor{black}{77.8$_{\smallsup{3.8}}$} & \textcolor{black}{73.5$_{\smallsup{2.8}}$} & \textcolor{black}{82.6$_{\smallsup{3.0}}$} & \textcolor{black}{72.8$_{\smallsup{7.9}}$} & \textcolor{black}{69.7$_{\smallsup{14.6}}$} \\
\hspace{10pt}\text{+ \sc{TA}} & \textcolor{black}{84.8$_{\smallsup{0.7}}$} & \textcolor{black}{64.6$_{\smallsup{6.3}}$} & \textcolor{black}{71.5$_{\smallsup{4.0}}$} & \textcolor{black}{85.5$_{\smallsup{1.4}}$} & \textcolor{black}{79.0$_{\smallsup{4.5}}$} & \textcolor{black}{38.5$_{\smallsup{3.0}}$} & \textcolor{black}{78.9$_{\smallsup{2.4}}$} & \textcolor{black}{81.2$_{\smallsup{3.9}}$} & \textcolor{black}{77.5$_{\smallsup{1.4}}$} & \textcolor{black}{88.6$_{\smallsup{1.3}}$} & \textcolor{black}{78.2$_{\smallsup{6.6}}$} & \textcolor{black}{77.0$_{\smallsup{6.3}}$} \\
\hspace{10pt}\text{+ \sc{ST}}
& \textcolor{black}{69.3$_{\smallsup{9.2}}$} &
\textcolor{black}{74.3$_{\smallsup{1.2}}$} &
\textcolor{black}{85.4$_{\smallsup{1.7}}$} &
\textcolor{black}{81.9$_{\smallsup{12.2}}$} &
\textcolor{black}{79.9$_{\smallsup{4.8}}$} &
\textcolor{black}{42.0$_{\smallsup{1.5}}$} &
\textcolor{black}{82.8$_{\smallsup{2.3}}$} &
\textcolor{black}{77.3$_{\smallsup{3.1}}$} & \textcolor{black}{73.1$_{\smallsup{2.3}}$} &
\textcolor{black}{88.1$_{\smallsup{1.3}}$} &
\textcolor{black}{81.2$_{\smallsup{0.5}}$} &
\textcolor{black}{53.9$_{\smallsup{4.3}}$} \\
\hspace{10pt}\text{+ \sc{ITFT$_{\smallsup{MNLI}}$} \text{+ ST}} & \textcolor{black}{85.4$_{\smallsup{0.3}}$} &
\textcolor{black}{74.8$_{\smallsup{0.7}}$} &
\textcolor{black}{86.1$_{\smallsup{1.1}}$} &
\textcolor{black}{89.7$_{\smallsup{0.7}}$} &
\textcolor{black}{86.2$_{\smallsup{4.2}}$} &
\textcolor{black}{42.2$_{\smallsup{2.0}}$} &
\textcolor{black}{84.1$_{\smallsup{1.7}}$} &
\textcolor{black}{84.3$_{\smallsup{2.0}}$} & \textcolor{black}{78.4$_{\smallsup{1.3}}$} &
\textcolor{black}{89.3$_{\smallsup{1.0}}$} &
\textcolor{black}{81.4$_{\smallsup{1.2}}$} &
\textcolor{black}{72.7$_{\smallsup{5.4}}$} \\
\hspace{10pt}\text{+ \lstrata} &
\textcolor{black}{\textbf{87.3}$_{\smallsup{0.3}}$} &
\textcolor{black}{\textbf{75.1}$_{\smallsup{0.2}}$} &
\textcolor{black}{\textbf{86.}4$_{\smallsup{0.8}}$} &
\textcolor{black}{\textbf{91.7}$_{\smallsup{0.7}}$} &
\textcolor{black}{\textbf{87.3}$_{\smallsup{2.9}}$} &
\textcolor{black}{\textbf{43.0}$_{\smallsup{2.3}}$} &
\textcolor{black}{\textbf{84.5}$_{\smallsup{1.6}}$} &
\textcolor{black}{\textbf{86.3}$_{\smallsup{1.8}}$} & \textcolor{black}{\textbf{79.0}$_{\smallsup{1.0}}$} &
\textcolor{black}{\textbf{90.0}$_{\smallsup{0.6}}$} &
\textcolor{black}{\textbf{81.5}$_{\smallsup{0.7}}$} &
\textcolor{black}{\textbf{77.1}$_{\smallsup{5.4}}$} \\
\bottomrule
\end{tabular}
\end{adjustbox}
\caption{Our experiment results in the \ssc{Few-shot} data regime.}
\label{tbl:8}
\end{table*}

\end{document}